\RequirePackage{fix-cm}
\documentclass[smallextended]{svjour3}       % onecolumn (second format)
\smartqed  % flush right qed marks, e.g. at end of proof
\usepackage{graphicx}

\usepackage{hyperref}
\usepackage{csquotes}
\usepackage[most]{tcolorbox}
\usepackage{natbib}

\usepackage{listings}
\usepackage{booktabs}
\usepackage{multirow}
\usepackage{enumitem}

\begin{document}

% not sure about this title; it's not very engaging and doesn't say much about the central problem / goal; I will try to suggest something after having read the paper
% UPDATE: here are some proposals for you to play around with
% How Are Collaboration and Communication Skills Mentioned in ML Job Postings? A Global LLM-Supported Analysis
% The Human Side of ML Engineering: Analyzing the Collaboration and Communication Skill Requirements in ML Job Postings
% Team Players Wanted: What ML Job Ads Say About Communication and Collaboration Skills
% Collaboration and Communication Skill Requirements in ML Job Postings: A Reusable Hybrid Human-LLM Methodology
\title{The Human Side of ML Engineering: Analyzing the Collaboration and Communication Skills in ML Job Postings}

\titlerunning{Analyzing the Collaboration and Communication Skills in ML Job Postings}        % if too long for running head

\author{Aidin Azamnouri \and Phuong Thao Le \and Justus Bogner \and Stefan Wagner
}


\institute{Aidin Azamnouri \and Phuong Thao Le \and  Stefan Wagner \at
            Chair of Software Engineering, TUM School of Computation, Information and Technology, Technical University of Munich, Heilbronn, Germany
              %\email{fauthor@example.com}
           \and
           Justus Bogner \at
              Software and Sustainability (S2) Group, Vrije Universiteit Amsterdam, Amsterdam, The Netherlands
}

\date{Received: date / Accepted: date}

\maketitle

\begin{abstract}
Studies of Machine Learning (ML) practice highlight the importance of combining technical expertise with interpersonal capabilities, particularly collaboration and communication. However, evidence on how ML job postings express these skills is scarce, and prior work largely relies on keyword-based analyses that overlook implicit expression. We analyze demand for these skills in ML job postings across regions, industries, and seniority levels, and compare it with software engineering (SE) roles. We collected 1,000 postings (500 ML, 500 SE) from five continents via a semi-automated pipeline. For annotation, we synthesized a taxonomy of collaboration and communication sub-categories from prior work and used an LLM as a third annotator alongside two humans, enabling detection of context-dependent and implicit skill references. Collaboration appeared in 88.6\% of ML postings and communication in 72.2\%, far above prior keyword-based estimates for AI roles. Roughly one in five collaboration and one in four communication requirements were expressed only implicitly. Decomposing these requirements into the taxonomy's sub-categories reveals what aggregate figures conceal. Collaboration is demanded predominantly as integration across role boundaries rather than team membership. Communication is mostly required in generic terms, with fewer than half of the postings requiring it naming whom the candidate must communicate with. ML and SE postings demand interpersonal competence at statistically indistinguishable levels but orient it differently, SE toward team practices and ML toward presentation and audience-facing interaction. Within ML, the advertised role separates interpersonal profiles more sharply than any other dimension, including the contrast with SE.

\end{abstract}

\section{Introduction}

Digital transformation is reshaping the competencies that define employability across industries~\citep{Cascio_2016, van_Laar_2017}. In technical domains such as software engineering (SE) and machine learning (ML), a growing body of research shows that professional success depends not on technical proficiency alone but also on the interpersonal capabilities required to operate within complex, interdisciplinary teams. Studies of ML practice in particular identify collaboration and communication as persistent challenges. Building production ML systems requires sustained coordination between data scientists, engineers, and domain experts, and breakdowns at these interfaces are a recurring source of project friction~\citep{Nahar_2022, Mailach_2023, Azamnouri_2025, Azamnouri_2026}. Translating technical artifacts into organizational value thus depends as much on communication and coordination as on algorithmic skill. While the importance of digital and technical hard skills is widely acknowledged, there is growing consensus that human agency, reasoning, and interpersonal abilities remain irreplaceable by automation~\citep{Pol_kov_2023}. Within information technology (IT), the rapid emergence of complex socio-technical systems has amplified the demand for professionals who combine technical competence with sophisticated non-technical abilities~\citep{Ternikov_2022}. \textbf{Whether and how this documented importance is reflected in the requirements employers actually articulate when hiring for ML roles, however, remains largely unexamined.}

The term soft skills encompasses the interpersonal, communicative, and self-regulatory abilities that allow individuals to coordinate effectively in a workplace setting~\citep{Sayfullina_2018}. Whereas technical or domain knowledge concerns what is being done, soft skills concern how work is carried out through behavior and interaction~\citep{Tamburri_2020}. Major occupational classification systems acknowledge this distinction by labeling such competencies \enquote{transversal skills} (ESCO\footnote{European Skills, Competences, Qualifications and Occupations classification, maintained by the European Commission: \url{https://esco.ec.europa.eu/}}) or \enquote{cross-functional skills} (O*NET\footnote{Occupational Information Network, sponsored by the U.S. Department of Labor: \url{https://www.onetonline.org/}}), in recognition of their applicability across heterogeneous tasks involving interaction, coordination, and complex decision-making.

This study focuses specifically on two soft skills that are widely regarded as foundational for technical work: collaboration and communication~\citep{Nahar_2022, Mailach_2023, Azamnouri_2025, Azamnouri_2026, Azamnouri_2026_b}. Collaboration denotes the capacity to work toward shared objectives by integrating diverse expertise and providing mutual support in solving complex problems, whereas communication refers to the exchange of meaning and intent through verbal, written, and non-verbal channels in service of coordination and shared understanding~\citep{Tripathy_2018, Tripathy_2020}. In the IT sector, these skills are not social pleasantries but empirically demonstrated drivers of project success, software quality, and developer productivity~\citep{Angelov_2016, Balcar_2016}. A growing share of practitioners now prefers candidates with strong interpersonal abilities and average technical skills over the inverse profile, reflecting the observation that the most pressing challenges in modern software projects are often human-centric rather than purely algorithmic~\citep{Kalliamvakou_2019}.

ML, a subset of artificial intelligence (AI) concerned with algorithms that learn from data and improve performance over time, has evolved from an academic niche into a foundational element of global industrial strategy~\citep{Pol_kov_2023}. ML systems are no longer confined to specialized technological functions but are deeply embedded in healthcare, finance, manufacturing, and the public sector. As these systems are integrated into mainstream operations, collaboration between ML engineers, data scientists, and non-technical stakeholders has become an essential feature of professional practice. Recent evidence indicates that adaptability, critical thinking, teamwork, and communication appear in nearly twenty percent of AI-related job postings~\citep{Vu_2026}, reflecting the interdisciplinary nature of ML work. Engineers and data scientists must regularly translate complex analytical findings into actionable insights for decision-makers without a technical background. As ML systems exert greater influence on societal outcomes, empathy, ethical awareness, and collaborative problem-solving have likewise shifted from supplementary attributes to core professional requirements for the responsible development and deployment of these systems.

Despite the apparent significance of these competencies, scientific literature provides little focused evidence on the concrete demands for collaboration and communication skills in ML roles specifically~\citep{Piorkowski_2021, Mailach_2023}. Existing job-posting analyses have largely concentrated on generic IT positions or traditional SE roles~\citep{Galster_2022, Pol_kov_2023}, and consequently do not account for regional differences or the unique socio-technical demands of the ML lifecycle. A second limitation is methodological. Prior work has relied predominantly on keyword-based extraction, which fails to capture the indirect ways in which recruiters express the need for soft skills. A job posting may not contain the word communication, yet may nonetheless require \enquote{the ability to explain solutions to non-technical partners}. Similarly, collaboration may surface only as a request for \enquote{a team player in an agile environment}. Detecting such implicit expressions at scale demands more sophisticated extraction techniques. Conversely, exhaustive manual annotation is costly, slow, and prone to inconsistency when applied to large multilingual corpora. To navigate this trade-off, the present study employs large language models (LLMs), specifically the LLM-as-a-Judge paradigm~\citep{Zheng_2023, Gu_2026} integrated with a human-in-the-loop annotation process, to analyze multilingual job-market data with a level of contextual sensitivity that keyword pipelines cannot achieve. Our findings suggest that this hybrid approach can substantially reduce the human labor required for large-scale text evaluation, though LLMs at present complement rather than replace human judgment.

Building on this motivation, the primary objective of this paper is to provide a comprehensive, data-driven analysis of the demand for collaboration and communication skills in ML job postings, considering both explicit and implicit expressions of these skills, and to contrast the observed patterns with those in traditional SE roles. We move beyond anecdotal accounts to deliver empirical evidence that can inform educators, recruiters, job seekers, and researchers.

Hence, the contributions of this paper are fourfold.
\begin{itemize}
    \item \textbf{For educators}, we quantify not only the prevalence of collaboration and communication requirements in ML hiring but their composition, identifying which specific interpersonal competencies employers name and how this composition shifts across seniority levels and advertised roles, providing an empirical basis on which universities can target curricular reform rather than adding generic soft-skill content.
    \item \textbf{For recruiters}, we offer a systematic taxonomy of collaboration and communication expressions together with evidence of a specificity gap. Requirements are asserted far more often than the audience, channel, or counterparty is named, which limits both what candidates can prepare for and what screening can assess.
    \item \textbf{For job seekers}, particularly those targeting senior or management-level positions, we identify which interpersonal competencies distinguish senior and management postings from earlier career stages.
    \item \textbf{For researchers}, we contribute a novel and scalable pipeline for job-market analysis that integrates LLM technology with established content-analysis techniques. The pipeline operationalizes an integrated human--LLM workflow for soft-skill classification, in which LLM-generated reasoning serves as the basis for incrementally refining model performance through prompt engineering, and provides a principled mechanism for surfacing implicit expressions of soft skills embedded in nuanced and indirect language.
\end{itemize}

The remainder of this paper is organized as follows. Section~\ref{sec:background-related-work} reviews related work on soft-skill analysis in software and ML labor markets and on LLM-based annotation methods. Section~\ref{sec:methodology} presents the study approach, data collection, translation, and annotation pipeline. Section~\ref{sec:results} reports the empirical findings on the prevalence and articulation of collaboration and communication skills across regions, industries, and seniority levels. Section~\ref{sec:discussion} discusses implications for stakeholders and reflects on the methodological contribution, before Section~\ref{sec:threats-to-validity} discusses the threats to validity. Section~\ref{sec:conclusion} concludes and outlines avenues for future work.

\section{Background and Related Work}
\label{sec:background-related-work}

We first review evidence on the labor-market demand for collaboration and communication skills, then narrow the lens to SE and ML roles, and finally summarize methodological work on job-posting collection, skill extraction, and LLM-based evaluation, before articulating the gap that motivates the present study.
 
\subsection{Collaboration and Communication in the Global Labor Market}

Collaboration denotes the ability to work with others toward shared objectives through mutual support, coordinated effort, and the integration of diverse expertise~\citep{Tripathy_2018, Tripathy_2020}; communication refers to the exchange of information, meaning, and intent through verbal, written, and non-verbal channels~\citep{Saxena_2022}. As work becomes increasingly digital and distributed, both competencies underpin coordination, shared understanding, and timely issue resolution within organizations~\citep{Tripathy_2020}.
 
Empirical studies consistently show that employers rank these competencies alongside technical qualifications. Employer surveys in Malaysia~\citep{Bee_2015} and Italy~\citep{Tamburri_2020} find that hiring managers prioritize communication and teamwork when evaluating candidates and graduates; longitudinal analyses document rising demand for collaboration, communication, and people-management skills even in traditionally technical sectors such as the U.S. energy industry~\citep{Lyu_2021} and Slovak IT and engineering~\citep{Pol_kov_2023}; and an integrative review identifies communication and teamwork as core employability skills across professions~\citep{Messum_2017}. Labor-economic evidence shows that this demand translates into measurable outcomes, with occupations combining strong cognitive and social skill requirements commanding higher wage premiums and faster employment growth~\citep{Deming_2018}. The pattern extends beyond STEM: among Japanese-language graduates, communication-intensive roles display the strongest alignment between skill competency and job performance~\citep{Setiana_2019}. Persistent gaps between employer expectations and graduate preparedness nevertheless remain~\citep{Zainal_Shah_2022}.
 
Because these findings replicate across countries, sectors, and methodologies, the centrality of collaboration and communication is unlikely to be an artifact of any single labor market or measurement approach. What this literature does not establish, however, is whether and how these dynamics manifest in ML roles specifically, whose interdisciplinary and stakeholder-facing character makes interpersonal competencies plausibly even more salient.
 
\subsection{Soft-Skill Demand in SE and AI Roles}
 
Software-related occupations mirror the broader labor market. Manually analyzing 530 software job advertisements in New Zealand,~\citet{Galster_2022} identified seventeen soft skills, with communication-related skills the most frequently requested across all position types.~\citet{Pol_kov_2023}, recoding 3,461 soft-skill terms from five years of Slovak postings (2017--2021) into sixteen Industry 5.0--relevant categories, found that soft skills accounted for more than 40\% of required competencies even in IT and engineering, with critical and analytical thinking, problem-solving, and communication the most consistently sought-after.
 
For AI-related roles, the evidence base is thinner and partly unvalidated. The most directly related effort, a large-scale analysis of AI postings available so far only as a preprint, reports that adaptability and critical thinking, collaboration, and communication constitute the dominant soft-skill clusters in AI roles~\citep{Vu_2026};~\citet{De_Morais_Le_a_2025} identify curiosity, critical thinking, empathy, and ethical awareness as essential competencies for data scientists working in interdisciplinary teams and communicating insights to non-technical stakeholders. While this evidence suggests that collaboration and communication are indispensable in both SE and AI roles, the only large-scale account of AI roles remains unreviewed, and ML positions are typically treated as a subset of broader IT or AI aggregates rather than analyzed in their own right, leaving open how these skills are concretely articulated in ML job postings.

Educational research has begun to respond to these demands with interdisciplinary, project-based course formats in which students from heterogeneous study programs build AI prototypes in mixed teams, with collaboration and communication treated as explicit learning objectives rather than by-products~\citep{Azamnouri_2026_c}. Such designs are, however, motivated by general appeals to the interdisciplinary nature of AI work rather than by evidence on which interpersonal competencies employers actually name, which is precisely what a compositional analysis of hiring requirements can supply.
 
\subsection{Collecting Job Postings for Labor-Market Analysis}
 
Strategies for collecting job postings have evolved from manual downloading and filtering from employer websites or job boards~\citep{Bee_2015, Kovacs_2022} to automated pipelines in which postings retrieved through web crawlers or APIs undergo standard NLP preprocessing~\citep{Muthyala_2017, Sioziou_2024}, complemented by public datasets such as those hosted on Kaggle~\citep{Sayfullina_2018, Green_2022} and by scheduled scraping systems that enable continuous, longitudinal acquisition across platforms~\citep{Mbah_2017}. The trajectory is clear: from manual, small-scale collection toward reproducible, large-scale aggregation pipelines that support timely analysis of skill trends.
 
\subsection{Methods for Skill Extraction}
\label{sec:methods-for-skill-extraction}
 
Methods for extracting skills from job postings have evolved along five broadly overlapping lines, which differ in how well they handle the implicit, context-dependent expressions of soft skills characteristic of ML postings.
 
\textbf{Keyword and dictionary-based methods.}~\citet{Kovacs_2022} analyzed over 5,000 digital-marketing postings using keyword-based coding, confirming the consistent emphasis on communication, teamwork, and interpersonal ability;~\citet{Verma_2019, Verma_2021} applied content analysis to advertisements across multiple domains, including AI and ML; and~\citet{Mahdi_2021} propose attention-based keyword extraction from job descriptions to support search-engine optimization and resume screening.
 
\textbf{Inductive and manual coding.} Other studies derive soft-skill categories inductively from the data.~\citet{Galster_2022} used bottom-up coding on 530 SE postings to identify seventeen context-specific soft skills, an approach that surfaces nuanced expressions but is labor-intensive and not readily scalable, and~\citet{Coetzee_2024} combine qualitative analysis with text mining to extract hard and soft skills from data analytics postings.
 
\textbf{Statistical and topic-modeling methods.} To scale beyond manual review,~\citet{Muthyala_2017} normalize, lemmatize, and vectorize postings into bag-of-words representations for skill-pattern identification,~\citet{Mbah_2017} built an Elasticsearch-based labor-market intelligence system supporting longitudinal trend analysis, and~\citet{Siswipraptini_2023} applied hierarchical clustering and semantic analysis to IT postings. These approaches scale well but rely largely on term-level co-occurrence rather than deeper semantic interpretation.
 
\textbf{Semantic and embedding-based methods.} TaxoSoft combines DBpedia knowledge graphs with Word2Vec embeddings trained on job-posting corpora to expand soft-skill vocabularies beyond exact keywords, capturing variation such as \enquote{team collaboration} versus \enquote{working well with others}~\citep{Khaouja_2019}. Related pipelines apply embeddings to labor-market analysis~\citep{Romanko_2022}, semi-automatic crowdsourced extraction~\citep{Calanca_2019}, large-scale job recommendation~\citep{Shalaby_2017}, and skill recommendation from job-title embeddings~\citep{Ong_2023}.
 
\textbf{Neural sequence labeling.} State-of-the-art extraction identifies skill mentions directly from text without predefined keyword lists. SkillSpan, a transformer-based span-extraction framework, uses contextual embeddings to determine whether a phrase functions as a skill in context, e.g., distinguishing \enquote{ability to collaborate across teams} from generic mentions of teamwork, and generalizes well across domains~\citep{Zhang_2022}.~\citet{Skondras_2023} similarly apply BERT and feed-forward networks to classify postings and extract skill spans, and a recent survey reports that transformer architectures consistently outperform dictionary- and embedding-based methods~\citep{Senger_2024}. Extracted spans are typically standardized through ontology alignment, e.g., ESCO, or embedding-similarity mapping.
 
\textbf{Summary.} Across these five lines, the literature progresses from rule-based coding toward increasingly context-aware models, yet a common limitation persists. Most methods identify isolated skill spans rather than interpret the broader contextual meaning of skill requirements within complete job descriptions. This limitation is particularly consequential for collaboration and communication in ML postings, which are frequently expressed indirectly through cross-functional team responsibilities, stakeholder-facing duties, or project-based workflows rather than as discrete keywords. Detecting such implicit expressions at scale requires methods that reason over the surrounding context, motivating evaluative LLM-based pipelines.
 
\subsection{LLMs as Evaluators in Annotation Pipelines}
 
Human evaluation remains the gold standard for text annotation but is costly, time-intensive, and difficult to scale~\citep{He_2025, Croxford_2025}, while traditional automated metrics such as BLEU, ROUGE, and Pass@k depend on reference outputs and fail to capture qualitative dimensions such as readability, coherence, and domain relevance~\citep{He_2025, Wang_2025}. The LLM-as-a-Judge paradigm addresses this gap by employing LLMs as evaluators capable of producing reasoned, multi-dimensional assessments.~\citet{Wang_2025} distinguish embedding-based, probability-based, and output-based variants, the last being the most representative because the model provides reasoning and justifications that approximate expert review. Empirical evidence reports near-human performance in selected tasks.~\citet{Croxford_2025} document a 96\% reduction in evaluation time relative to human experts in clinical artifact assessment, and~\citet{Ahmed_2025} find that LLMs can replace 9--33\% of manual annotation effort in SE without statistically significant changes to inter-rater agreement, provided model confidence is used to triage samples.
 
The paradigm nevertheless carries known risks: task-dependent reliability, reduced consistency on tasks requiring nuanced interpretation, and susceptibility to verbosity, position, and egocentric biases~\citep{Wu_2025, Wang_2025}, as well as hallucinated judgments drawn from pre-training associations rather than the artifact at hand and polarized scores from smaller models~\citep{Croxford_2025}. Reported mitigations, including role-playing prompt designs~\citep{Wu_2025}, reasoning-focused architectures~\citep{Croxford_2025}, and human-in-the-loop review of low-confidence cases~\citep{He_2025}, converge on a methodological precedent that our pipeline adopts. LLM evaluators are most reliable when paired with human annotators and grounded in carefully engineered prompts.
 
\subsection{Research Gap}
 
Despite the growing recognition of soft skills in technical domains, no prior study offers a focused empirical analysis of collaboration and communication skills in ML job postings specifically. ML positions are typically subsumed within broader IT or AI aggregates, e.g.,~\citep{Vu_2026}, and the few studies addressing ML-adjacent roles concentrate on data scientists rather than ML engineers and treat soft skills as high-level themes rather than objects of systematic measurement~\citep{De_Morais_Le_a_2025}.
 
A second gap is methodological. Existing skill-extraction approaches, from keyword matching to neural sequence labeling, remain oriented toward isolated skill spans rather than the broader contextual meaning of skill requirements. This orientation is poorly suited to collaboration and communication in ML postings, which are frequently embedded in descriptions of cross-functional collaboration, stakeholder engagement, and team-based workflows rather than stated as discrete keywords. While LLM-based evaluation pipelines have begun to demonstrate the contextual reasoning required for such judgments, their application to labor-market research remains emerging and has yet to be integrated with structured human annotation in a hybrid workflow.
 
The present study addresses both gaps. Empirically, we provide the first dedicated cross-regional analysis of how collaboration and communication skills are presented in ML job postings, with a comparison against SE as a reference profession. Methodologically, we combine traditional content analysis with an LLM-as-a-Judge annotation architecture in which an LLM serves as a third annotator alongside two humans, enabling the detection of both explicit and implicit expressions of soft skills at scale.

\section{Methodology}
\label{sec:methodology}

\begin{figure}
\centering
\includegraphics[width=0.7\textwidth]{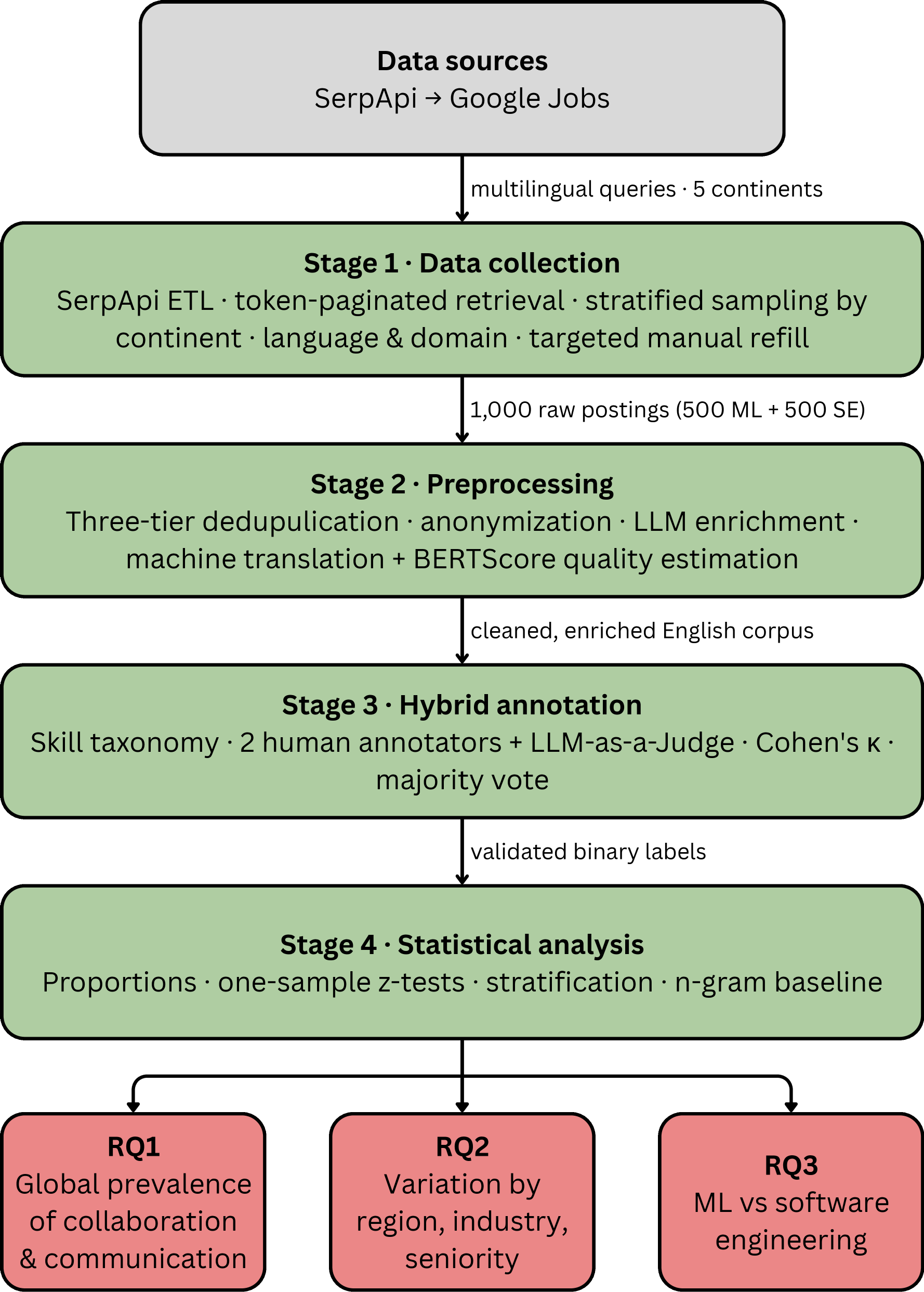}
\caption{An overview of the study process}
\label{study-process}
\end{figure}

This study follows a four-stage empirical pipeline, depicted in Fig.~\ref{study-process}. Job postings were collected through the SerpApi interface to Google Jobs and supplemented with targeted manual scraping. The raw corpus was preprocessed through cleaning, anonymization, feature enrichment, and machine translation with automated quality estimation. The cleaned corpus was then annotated using a hybrid protocol that combined two human annotators with an LLM-as-a-Judge serving as a third independent annotator. Finally, the validated annotations were analyzed through descriptive and inferential statistical methods to address the three research questions. Each stage is described in detail below.

\subsection{Research Questions}

To operationalize the research objectives, this study is organized around three research questions that move progressively from prevalence to contextual variation to cross-domain comparison.

\textbf{RQ1:} To what extent are collaboration and communication skills required in ML job postings?

This question quantifies the prevalence of the two target skills across the ML labor market and tests whether they constitute core requirements rather than peripheral attributes.

\textbf{RQ2:} To what degree does the demand for collaboration and communication skills in ML roles vary across geographic regions, industry domains, and seniority levels?

This question disaggregates the global picture along three contextual dimensions:
\begin{enumerate}[label=(\roman*)]
    \item \textit{Region.} Whether geographic context (e.g., Europe, North America, Asia) shapes the demand for collaboration and communication in ML roles.
    \item \textit{Industry.} Whether the demand differs across industry domains such as healthcare, industrial, technology, and finance.
    \item \textit{Seniority.} Whether the importance of collaboration and communication varies across seniority levels, from entry-level through mid, senior, and managerial roles.
\end{enumerate}

\textbf{RQ3:} To what degree do the demands for collaboration and communication skills differ between ML and software engineering job postings?

This comparative question evaluates whether the data-driven, stakeholder-facing nature of ML work elicits a different level of interpersonal demand than is observed in traditional SE positions.

To answer these questions, the study implements an end-to-end empirical pipeline encompassing data collection, preprocessing, annotation, and statistical analysis. In contrast to the manual collection strategies prevalent in earlier work, postings were retrieved through an automated pipeline built on SerpApi\footnote{\url{https://serpapi.com/}}, a regulated third-party interface to Google Jobs that aggregates listings from multiple recruitment platforms (e.g., LinkedIn, Glassdoor, and Indeed) through a single compliant gateway, supplemented by targeted manual sourcing for replacement of duplicates, yielding broader coverage at lower cost.
The semi-automated procedure produced a stratified, multi-country sample of 1,000 unique postings divided equally between the ML and SE domains and distributed across five continents (Asia, Europe, Africa, North America, and South America). Non-English postings were translated using a multi-model machine-translation pipeline, whose output was automatically evaluated with BERTScore to ensure semantic fidelity, and additionally validated through manual review of the German and Vietnamese subsets by annotators who were native or fluent speakers of those languages.

A central methodological contribution is a hybrid annotation architecture in which an LLM-as-a-Judge serves as a third independent annotator alongside two human annotators, with task-specific LLMs classifying collaboration and communication skills. The LLM improved the annotation outcome in two concrete ways. It recovered implicit skill expressions that keyword-based detection misses, and its mandatory evidence-citing rationales served as an audit trail that surfaced evidence overlooked by the human annotators during disagreement review. The full three-way annotation conducted here doubles as a validation study for lighter-weight reuse. Because the LLM's labels closely track a high-agreement human ground truth, future applications of the pipeline can restrict human effort to LLM-flagged borderline cases and periodic audits rather than exhaustive dual annotation, which is what makes the framework scalable to longitudinal labor-market monitoring.

\subsection{Data Collection}
\label{sec:data-collection}

\begin{figure}
\centering
\includegraphics[width=1\textwidth]{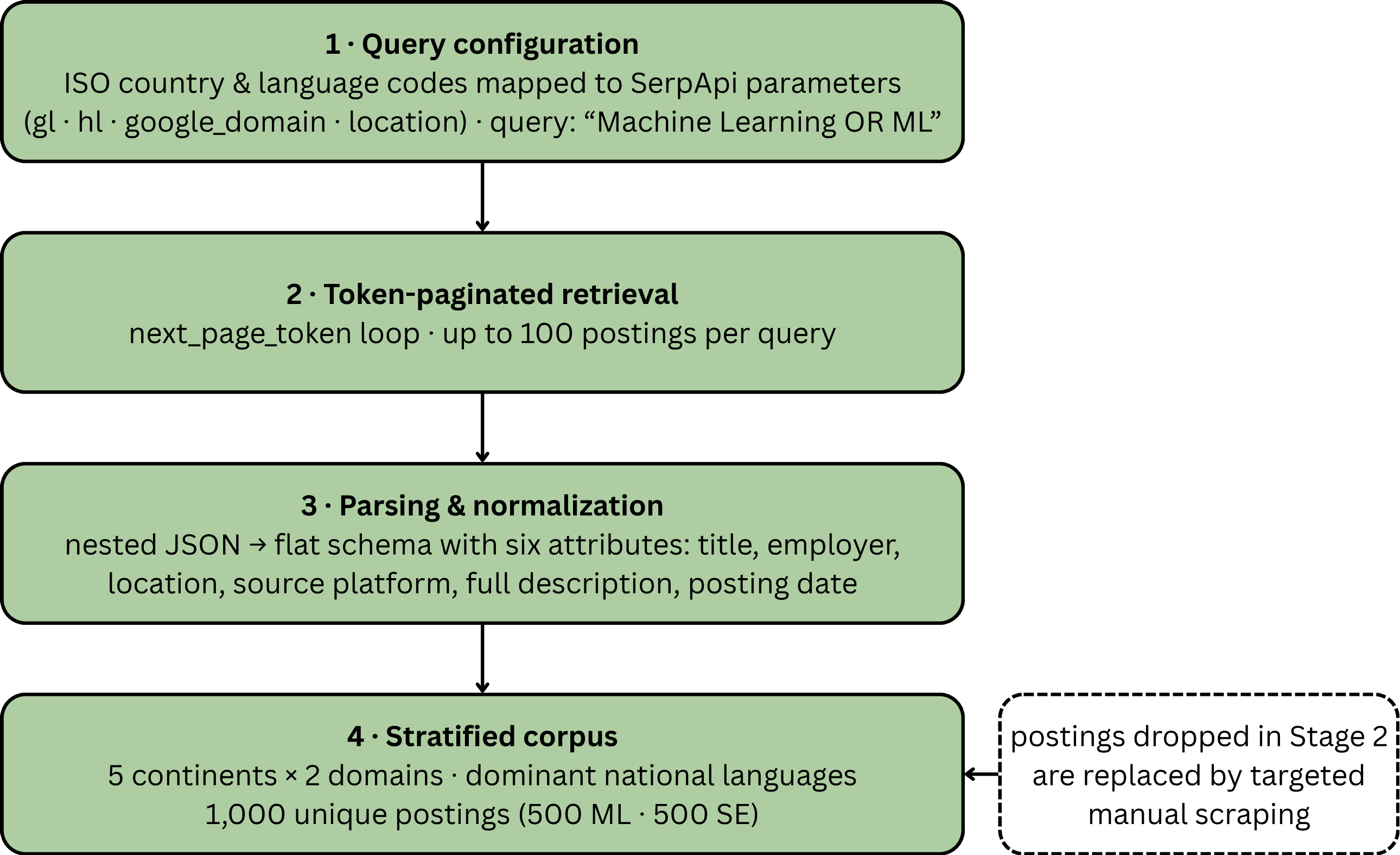}
\caption{An overview of the data collection stage}
\label{stage1}
\end{figure}

A central design decision concerned the means of data acquisition. Much of the prior literature relies on web-scraping frameworks such as Selenium or BeautifulSoup applied directly to platforms such as LinkedIn or Indeed; these approaches must navigate complex ethical and legal constraints imposed by platform Terms of Service. To ensure strict compliance with platform policies and data privacy norms, we instead used SerpApi. An overview of the data collection stage is presented in Fig.~\ref{stage1}.

Queries were configured by mapping ISO country and language codes to SerpApi's \texttt{gl}, \texttt{hl}, \texttt{google\_domain}, and \texttt{location} parameters within a custom Python script. As an illustration, the configuration in Listing~\ref{lst1} retrieves postings in Spanish that mention \enquote{Machine Learning} or \enquote{ML} and are issued from Colombia.

\begin{lstlisting}[caption={An example of SerpApi query.},label={lst1}]
{
  "engine": "google_jobs",
  "q": "Machine Learning OR ML",
  "location": "Colombia",
  "google_domain": "google.co",
  "gl": "co",
  "hl": "es",
  "api_key": API_KEY
}
\end{lstlisting}

A custom Python ETL (Extract, Transform, Load) pipeline orchestrated the extraction. To retrieve up to 100 postings per query, the script implemented a token-based pagination loop in which the \texttt{next\_page\_token} was updated on each iteration, allowing traversal of Google Jobs' indexed results without duplication or loss of continuity. For each posting, the script parsed the nested JSON response into a flattened schema and retained six core attributes: job title, hiring entity, geographic location, originating platform (the \texttt{via} field), full-text job description, and posting date. The posting date was extracted from the \texttt{detected\_extensions} metadata rather than from raw description text, providing a more reliable temporal marker. This procedure transformed volatile web listings into a normalized dataset suitable for subsequent natural language processing (NLP)-based analysis.

The final corpus comprises 1,000 unique postings, divided equally between the ML and SE domains. Because Google Jobs scopes queries to countries rather than continents, the stratified design was operationalized at the country level. Each of the five continents (Asia, Europe, Africa, North America, and South America) contributed a fixed quota of 100 postings per domain, drawn from countries with sizable ML/IT labor markets. Query language followed from the target country. Postings were retrieved in the dominant national language for Germany, France, Japan, Vietnam, Brazil, and Colombia (German, French, Japanese, Vietnamese, Portuguese, and Spanish), capturing localized hiring conventions that English-only sampling would obscure, and in English for anglophone and English-dominant markets, including the United States, Canada, the United Kingdom, South Africa, Nigeria, the United Arab Emirates, and India; local-language markets account for roughly two-fifths of the corpus, and only non-English postings were routed to the translation pipeline. Quotas were filled by paginating through Google Jobs' relevance-ranked results until 100 unique, relevant postings per stratum survived cleaning, with attrition compensated through targeted manual sourcing on Glassdoor and Indeed under the same compliance standards. Because the ranking algorithm is opaque, the sample constitutes a quota-based draw of highly ranked postings rather than a random draw of the underlying population.

\subsection{Data Preprocessing}

\begin{figure}
\centering
\includegraphics[width=1\textwidth]{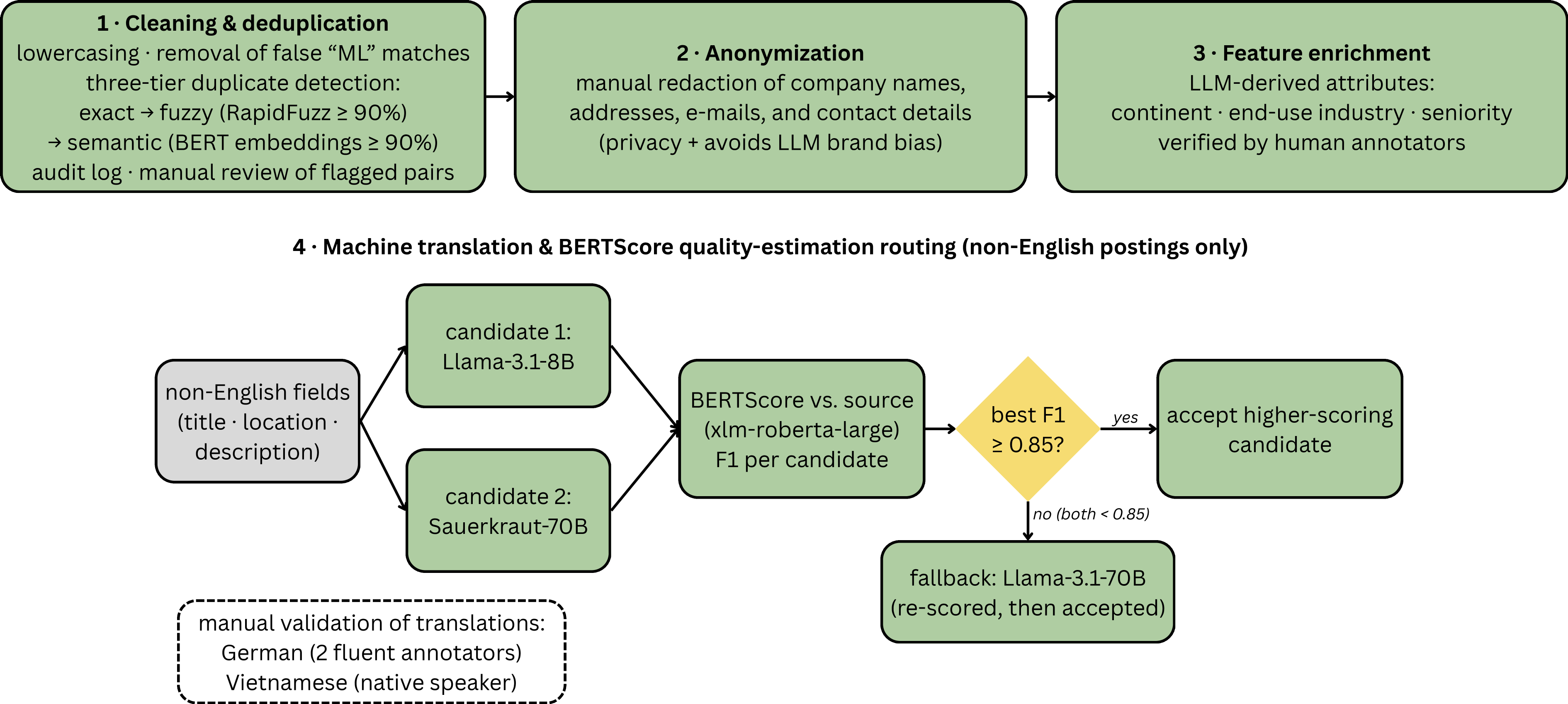}
\caption{An overview of the data preprocessing stage}
\label{stage2}
\end{figure}

The preprocessing pipeline, as shown in Fig.~\ref{stage2}, applies four operations, cleaning, anonymization, feature enrichment, and translation, to prepare the corpus for annotation while preserving semantic integrity.

\subsubsection{Cleaning and Deduplication}

All text was normalized to lowercase. Two recurring data-quality issues required attention. First, some postings retrieved by the specified query, e.g., Listing~\ref{lst1}, contained the token \enquote{ML} without referring to machine learning. Second, the SerpApi aggregation surface produces duplicates when the same posting is indexed across multiple platforms; such duplicates may be byte-identical or differ in superficial details.

To address duplication, we applied a three-tier detection procedure. Exact-match duplicates were removed using the duplicate-detection function. Lexical near-duplicates were detected with the rapid-fuzz library, which captures typographical variation and minor surface differences. Semantic near-duplicates were identified using BERT-based contextual embeddings, which capture paraphrase and conceptual equivalence beyond lexical overlap. A 90\% similarity threshold was applied for both fuzzy and semantic matching. For each detected near-duplicate pair, the line numbers, similarity score, and triggering text spans were logged to an audit spreadsheet, after which a manual review confirmed the removal decisions; flagged pairs identified as distinct vacancies sharing recruiter boilerplate were retained. A further 39 duplicates that had escaped automated detection (31 ML, 8 SE, about 4\% of the corpus) were identified by the annotators during the labeling phase, removed, and replaced through targeted manual sourcing on Glassdoor and Indeed to maintain the regional and domain stratification.

Anonymization served two purposes. First, downstream LLM-based classification could be biased toward associating prominent corporations, e.g., Google, Microsoft, Apple, with strong soft-skill cultures regardless of the content they post. Second, removing private information respects the privacy expectations of hiring entities. Company names, addresses, hiring emails, and other contact details were therefore redacted manually before the labeling phase.

\subsubsection{Feature Enrichment}

After cleaning and anonymization, LLMs were used to enrich the corpus with three categorical attributes used in subsequent analyses. The continent was derived from the location field (Asia, Africa, South America, North America, Europe). Industry domains (Industrial, Healthcare, Finance, or Technology) were inferred from the company description embedded in the job posting. Seniority level (Manager, Senior, Mid, Entry) was inferred jointly from the job title and description. All categorizations were verified by human annotators during the labeling phase.

\subsubsection{Multilingual Translation with Automated Quality Estimation}

Because roughly two-fifths of the corpus was collected in six non-English languages, a translation step preceded English-language annotation; human translation of the full multilingual portion was infeasible within the resources of this study. We therefore implemented a three-model translation pipeline with automated quality estimation. For each non-English field (title, location, description), two candidate translations are generated in parallel by Llama-3.1-8B-Instruct\footnote{\url{https://huggingface.co/meta-llama/Llama-3.1-8B-Instruct}} and Llama-3.1-SauerkrautLM-70B-Instruct\footnote{\url{https://huggingface.co/VAGOsolutions/Llama-3.1-SauerkrautLM-70b-Instruct}} (hereafter 8B and Sauerkraut) and scored against the source using BERTScore with xlm-roberta-large\footnote{\url{https://huggingface.co/FacebookAI/xlm-roberta-large}} as the multilingual encoder. BERTScore was preferred over n-gram metrics such as BLEU and METEOR because it operates on contextual embeddings and correlates highly with human judgments~\citep{Zhang_2020}. The router accepts the cheapest candidate whose F1 exceeds a fidelity threshold of 0.85, breaking ties in favor of 8B; only when both candidates fall below the threshold is Llama-3.1-70B-Instruct\footnote{\url{https://huggingface.co/meta-llama/Llama-3.1-70B-Instruct}} invoked as a fallback. The design objective is thus a guaranteed minimum semantic fidelity on every accepted translation at minimal inference cost, not a maximal BERTScore.
 
Table~\ref{tab:translation-routing} in the appendix summarizes translation quality and routing outcomes by language group and field type. Three regimes emerge. For typologically distant languages (Vietnamese, Japanese), noisy fields such as location frequently fail the threshold for both primary models, so the fallback handles the majority of these cases (53.8\%). For Portuguese/Spanish and German/French, routing is efficiency-driven. 8B suffices for short, formulaic fields, Sauerkraut is preferred for longer descriptions, where it offers a higher mean score and lower variance, and the fallback is rarely required (5.1\% and 8.0\% of location instances, respectively). Across all groups, a consistent text-type hierarchy holds. Titles are handled well by small models; locations expose threshold dynamics and data noise; and descriptions benefit from greater capacity. Because selection is governed by per-instance sufficiency rather than average performance, a model with a lower mean BERTScore can nevertheless be selected most frequently; for Portuguese/Spanish location strings, for example, Sauerkraut won 47 of 50 head-to-head comparisons, yet 8B was routed in 48.0\% of cases because its translations already cleared the fidelity floor.
 
To complement the automated quality estimation, all Vietnamese postings were manually reviewed by a native speaker, and all German postings by two annotators fluent in German and English; the reviews confirmed that the accepted translations preserved the meaning and contextual nuance of the originals. No comparable review was possible for the remaining languages; we discuss the residual risk in Section~\ref{sec:threats-to-validity}.

\subsection{Data Labeling}

\begin{figure}
\centering
\includegraphics[width=1\textwidth]{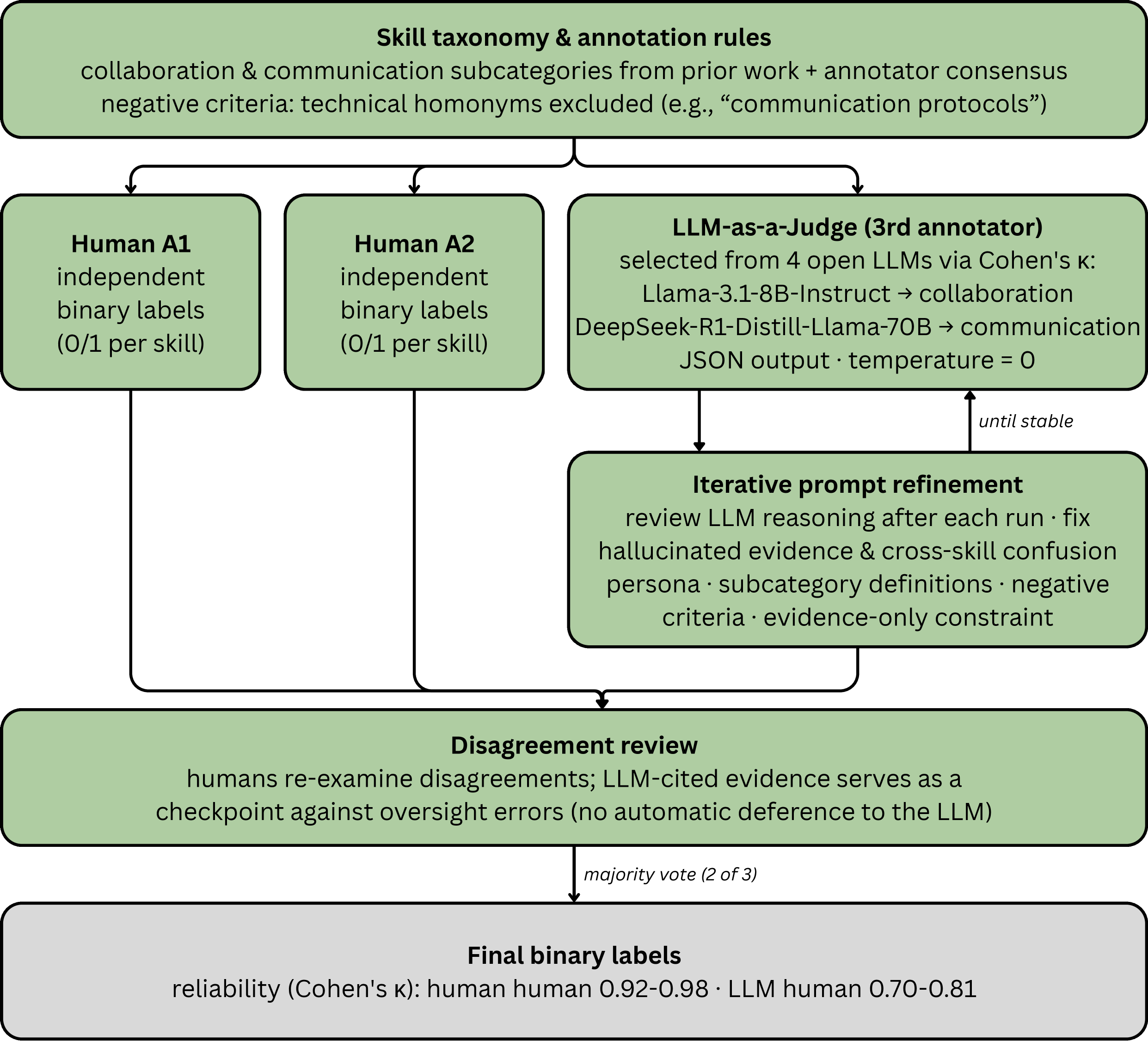}
\caption{An overview of the data labeling stage}
\label{stage3}
\end{figure}

The data labeling stage is illustrated in Fig.~\ref{stage3} and is discussed in detail below.

\subsubsection{Skill Taxonomy}
\label{sec:skill-taxonomy}

A soft-skill taxonomy was constructed by synthesizing prior findings. For instance, ~\citet{Pol_kov_2023} characterize communication as encompassing presentation, negotiation, client communication, active listening, and conflict resolution in both verbal and written modes.~\citet{Khaouja_2019} propose a taxonomy that incorporates stemmed surface forms, including verbs, adjectives, and adverbs, identifying terms such as interactive, communicator, interpersonal, communicate, and interaction for communication, and team, collaboration, cooperate, team player, and teamwork for collaboration. Building on these foundations and on the gold standard established through agreement between two human annotators, the operational categories used in this study were defined as follows.

\textbf{Collaboration:}
\begin{itemize}
    \item General collaboration (collaborate, collaborative, collaboration skills)
    \item Teamwork (team, development team, agile team, working with other team members)
    \item Cross-functional collaboration (cross-functional teams)
    \item Stakeholder coordination
\end{itemize}

\textbf{Communication:}
\begin{itemize}
    \item General communication (communication skills, strong communication, active listening)
    \item Interpersonal interaction (clients, stakeholders, customer relationships, interaction)
    \item Presentation and explanation (report, present, communicate complex technical ideas)
    \item Written communication (documentation, written reports)
    \item Conflict management
    \item Language proficiency
\end{itemize}

In addition, job postings that explicitly require leadership or that use job titles such as \enquote{leader} or \enquote{manager} were treated as requiring both collaboration and communication, on the grounds that prior work has demonstrated these competencies to be indispensable for managerial roles~\citep{Burke_2006, Morgeson_2009, Kalliamvakou_2019}. Leaders rely on communication to represent and motivate their teams~\citep{Olivia_Marbun_2023, Erbay_2024}, and effective leadership fuels collaboration through active listening and concern for team-member problems~\citep{Calvert_2018}. Mentorship was also considered evidence of communication when articulated as knowledge transfer and explanation rather than as a generic responsibility.

A further exclusion criterion concerns domain-technical homonyms. In ML and SE postings, the surface forms associated with the two target skills frequently appear in purely technical contexts in which they denote system properties rather than interpersonal competencies. Typical examples include \enquote{knowledge of API design, RESTful services, and communication protocols for AI/ML systems}, \enquote{experience optimizing networks for GPU-to-GPU communication}, and \enquote{investigating and debugging technical issues to ensure seamless communication between hardware and software}. Although each phrase contains the keyword \enquote{communication}, it refers to data exchange between software or hardware components and carries no implication about a candidate's interpersonal abilities. Analogous cases arise for collaboration, e.g., mentions of collaborative filtering algorithms or collaboration platforms as product features. Annotators, therefore, assigned a positive label only when the expression referred to human-to-human interaction; machine-to-machine and system-level usages were treated as negative evidence, and the same rule was encoded as a negative criterion in the LLM prompt. This requirement for contextual disambiguation further illustrates why naive keyword matching or underspecified LLM prompting is unsuitable for the task.

\subsubsection{Hybrid Human–LLM Annotation Protocol}

Two human annotators first agreed on the taxonomy described above and then independently labeled each posting to improve the validity of the final annotations. Binary labels were assigned per skill: 0 if the target skill was not identified and 1 if it was present. To complement human annotation and capture context-dependent or implicit expressions, an LLM-as-a-Judge served as a third independent annotator. The final ground-truth label for each posting was determined by majority vote across the three annotators.

Inter-annotator reliability was quantified using Cohen's $\kappa$~\citep{McHugh_2012}. Because Cohen's $\kappa$ is defined for agreement between two raters, it was applied pairwise. Separate coefficients were computed for each of the three annotator pairs (A1 vs. A2, A1 vs. LLM, and A2 vs. LLM). We deliberately report pairwise coefficients rather than a single multi-rater statistic such as Fleiss' $\kappa$ because the three annotators are not interchangeable and the pairings serve distinct evaluative purposes. Human-human agreement establishes the stability of the coding scheme and the upper bound of achievable reliability, while the two human-LLM coefficients assess whether the model reproduces human judgments and aligns comparably with both annotators rather than being biased toward one annotator's interpretive style. A pooled multi-rater coefficient would average over  this asymmetry. Cohen's $\kappa$ was preferred over alternatives such as Scott's $\pi$, Aickin's $\alpha$, and simple percent agreement because it offers a robust and interpretable estimate under marginal heterogeneity and corrects for chance agreement, which can be substantial in imbalanced binary tasks. The pairwise coefficients served only to assess annotation reliability; the final ground-truth label for each posting was consolidated separately, by majority vote across the three annotators.

\subsubsection{LLM Selection}

Four open-weight LLMs were evaluated for the third-annotator role: Llama-3.1-8B-Instruct, DeepSeek-R1-Distill-Llama-70B, Qwen3-30B, and Mistral-Large. These models were chosen for their strong reported performance on text understanding, reasoning, and classification, their open-weight availability, and their favorable trade-off between performance and computational cost~\citep{Grattafiori_2024, Guo_2025, Yang_2025, MistralAI_2024}. Each model was applied to the full ML subset, and its agreement with each human annotator was computed for both target labels.

Table~\ref{tab1} reports the resulting Cohen's $\kappa$ values. Human–human agreement is consistently very high (Cohen's $\kappa = 0.966$ for collaboration, Cohen's $\kappa = 0.913$ for communication), establishing a stable ground truth. Performance across LLMs, however, is task-dependent. For communication, DeepSeek achieves the highest agreement with both human annotators ($\kappa \approx 0.70$), outperforming the remaining models. For collaboration, Llama-3.1 dominates ($\kappa \approx 0.72-0.75$), while the other three models reach only fair-to-moderate agreement ($\kappa \approx 0.40-0.49$). On the basis of these results, a hybrid configuration was adopted in which Llama labels collaboration and DeepSeek labels communication, exploiting each model's relative strength.

\begin{table}
\caption{Cohen's $\kappa$ between human annotators (A1, A2) and each candidate LLM on the ML subset.}
\label{tab1}
\centering
\begin{tabular}{llcccc}
\toprule
 & & \multicolumn{4}{c}{LLM} \\
\cmidrule(lr){3-6}
Label & Annotator pair & Llama-3.1 & DeepSeek & Qwen3 & Mistral \\
\midrule
\multirow{3}{*}{collaboration}
  & A1 vs. A2  & \multicolumn{4}{c}{0.966} \\
  & A1 vs. LLM & 0.724 & 0.468 & 0.438 & 0.406 \\
  & A2 vs. LLM & 0.753 & 0.487 & 0.457 & 0.411 \\
\midrule
\multirow{3}{*}{communication}
  & A1 vs. A2  & \multicolumn{4}{c}{0.913} \\
  & A1 vs. LLM & 0.422 & 0.702 & 0.672 & 0.652 \\
  & A2 vs. LLM & 0.423 & 0.677 & 0.657 & 0.650 \\
\bottomrule
\end{tabular}
\end{table}

The LLMs employed in this study are specifically designed to comply with stringent data protection requirements by ensuring that sensitive conversational data is neither retained nor transmitted to any external servers.

\subsubsection{Iterative Prompt Refinement}

After the hybrid configuration was resolved, the prompt design was iteratively refined. After each annotation run, the model-generated explanations were reviewed alongside the labels. This human-in-the-loop validation surfaced two recurring failure modes. Hallucinated reasoning, in which justifications referenced phrases not present in the source posting, was the first. Cross-skill confusion, in which evidence for one skill was incorrectly used to justify a label for the other, was the second; for example, explicit references to teamwork were sometimes mislabeled as evidence of communication and vice versa, producing false positives that reduced precision.

To mitigate these failures, the prompt was progressively revised with clearer category definitions, stricter classification rules, and explicit negative criteria. After each run, misclassified examples were analyzed to identify the source of error, and the prompt was adjusted accordingly. Five design choices proved decisive in the final prompt.

First, the model is assigned the persona of a ~\enquote{labor market researcher,} which encourages a more analytical and cautious style of reasoning than open-ended classification prompts elicit. Second, each target skill is defined with explicit subcategories, e.g., stakeholder interaction, written communication, and cross-functional teamwork, removing reliance on the model's implicit understanding of these terms. Third, negative prompting is used to prevent over-interpretation. The model is explicitly instructed that environmental descriptions such as \enquote{we are a small team of engineers} are not sufficient evidence of collaboration because they describe a workplace rather than a required competency, and that phrases such as \enquote{partner with data scientists} indicate collaboration rather than communication, and that technical usages such as \enquote{communication protocols} or \enquote{GPU-to-GPU communication} denote machine-to-machine data exchange and must not be counted as evidence of communication skills. Fourth, the model is constrained to use only information explicitly stated in the posting, mitigating the LLM tendency to fill in plausible but unsupported context. Fifth, output is constrained to a fixed JSON schema, and the temperature is set to zero, ensuring deterministic and programmatically tractable responses. The complete prompt templates for both classification tasks, including all category definitions, negative criteria, and the required JSON output schema, are provided in our replication package\footnote{\url{https://doi.org/10.5281/zenodo.24205893}}.

System engineering decisions complemented the prompt design. Collaboration and communication classifications for each posting were dispatched in parallel across two threads, under a global rate limiter that spaced out API requests to match provider limits. When rate-limit errors occurred, the system honored server-provided retry hints when available and otherwise applied exponential backoff. Outputs were saved incrementally so the pipeline could resume after an interruption without losing progress, and a fallback parser was applied when responses contained extra text alongside the required JSON.

After the LLM labels and reasoning were produced, the human annotators reviewed cases of disagreement. Where the LLM cited evidence that genuinely supported a label different from the human annotation, the annotators revisited their initial labels to verify whether the evidence had been overlooked. Such corrections did occur, in both directions. In several cases, the LLM's mandatory evidence citation surfaced phrases the annotators had missed. In a posting for a senior engineering-management role, an initially negative communication label was revised after the LLM pointed to \enquote{mentoring} among the role's talent-management responsibilities and \enquote{coaching others} in its competency list; in a DevOps posting, the annotator had captured the explicitly stated communication requirement but initially overlooked the collaboration evidence, revising the label after the LLM cited \enquote{a good team player} and development \enquote{within a distributed software development environment with other contributors.} Conversely, the LLM's reasoning also corrected cross-skill confusion. One posting had initially been labeled positive for communication based on teamwork-related phrases, and the label was revised to negative after the LLM's explanation noted that, under the coding criteria, such phrases constitute evidence of collaboration rather than communication. The purpose of this review was not to defer to the LLM; in other disagreement cases, the annotators upheld their labels against the model, most often when it had treated a single unelaborated mention of mentoring as sufficient evidence of communication (see Section~\ref{sec:llm-limitations}). Rather, the LLM's reasoning served as an additional checkpoint that corrected human oversight errors and improved overall consistency.

\subsubsection{Final Annotation Reliability}

To establish that the annotation framework generalizes beyond ML and to enable the cross-domain comparison, the entire labeling protocol was replicated on the 500 SE job postings. Inter-human agreement is very high for both targets (Cohen's $\kappa = 0.975$ for collaboration, Cohen's $\kappa = 0.972$ for communication), reaffirming that the coding scheme operationalizes both constructs with sufficient clarity to be applied consistently by independent annotators. LLM–human agreement falls within the substantial range, with $\kappa = 0.712$ and $\kappa = 0.717$ for collaboration and $\kappa = 0.696$ and $\kappa = 0.702$ for communication. The pattern parallels the ML results in shape: human–human agreement is the upper bound, LLM–human agreement is consistently substantial, and the LLM aligns with both annotators to a comparable degree, indicating that the model is not biased toward either annotator's interpretive style.

After validation, the final inter-annotator agreement was computed for both the ML and SE subsets. On the ML subset, human–human agreement is almost perfect (Cohen's $\kappa = 0.980$ for collaboration, Cohen's $\kappa = 0.919$ for communication) as shown in Table~\ref{tab2}. LLM–human agreement is substantial across both labels (Cohen's $\kappa = 0.808-0.811$ for collaboration, Cohen's $\kappa = 0.754-0.804$ for communication), indicating that the model largely reproduces human judgments while differing in borderline cases. The SE subset shows a similar pattern as depicted in Table~\ref{tab3}. Human–human agreement remains very high ($\kappa = 0.975$ and $\kappa = 0.972$), and LLM–human agreement falls within the substantial range ($\kappa\approx0.70-0.72$). Across both subsets, the agreement levels confirm that the annotation framework is sufficiently stable and that the LLM output aligns reliably, if slightly more conservatively, with human judgment, supporting the use of the resulting labels for downstream analysis. Final labels were consolidated by majority vote across the three annotators.

\begin{table}
\caption{Final inter-annotator agreement by Cohen's $\kappa$ between human annotators (A1, A2) and the chosen LLM on the ML subset.}
\label{tab2}
\centering
\begin{tabular}{lccc}
\hline\noalign{\smallskip}
Label & A1 vs. A2 & A1 vs. LLM & A2 vs. LLM  \\
\noalign{\smallskip}\hline\noalign{\smallskip}
collaboration & 0.980 & 0.811 & 0.808 \\
communication & 0.919 & 0.804 & 0.754 \\
\noalign{\smallskip}\hline
\end{tabular}
\end{table}
 
\begin{table}
\caption{Final inter-annotator agreement by Cohen's $\kappa$ between human annotators (A1, A2) and the chosen LLM on the SE subset.}
\label{tab3}
\centering
\begin{tabular}{lccc}
\hline\noalign{\smallskip}
Label & A1 vs. A2 & A1 vs. LLM & A2 vs. LLM  \\
\noalign{\smallskip}\hline\noalign{\smallskip}
collaboration & 0.975 & 0.712 & 0.717 \\
communication & 0.972 & 0.696 & 0.702 \\
\noalign{\smallskip}\hline
\end{tabular}
\end{table}

\subsection{Facet-Level Coding of Skill Expressions}
\label{sec:facet-coding}

The binary labels established above answer whether a posting requires collaboration or communication, but not through which kind of expression the requirement is articulated. Because the taxonomy of Section~\ref{sec:skill-taxonomy} already decomposes both constructs into sub-categories, we added a second coding pass that attributes every positively labeled posting to one or more of these sub-categories, which we refer to as facets. Facet coding is an attribution rather than a detection task. It operates only on postings that the hybrid human--LLM pipeline has already labeled positive, and it does not alter any label.

\textbf{Facet inventory.} We operationalize five collaboration facets (general collaboration, cross-functional collaboration, teamwork and team practices, stakeholder coordination, team leadership) and seven communication facets (general communication, audience-facing interaction, presentation and explanation, mentoring and knowledge transfer, written communication, language proficiency, conflict and negotiation). These follow the taxonomy of Section~\ref{sec:skill-taxonomy} with three adjustments. First, the taxonomy's collaboration sub-category \enquote{stakeholder coordination} and its communication sub-category \enquote{interpersonal interaction} refer to the same counterparties and are separable only by what the posting asks the candidate to do with them; we therefore assign a mention to the collaboration facet when it is governed by a coordination verb (\enquote{work with}, \enquote{partner with}, \enquote{coordinate}, \enquote{align with}) and to the communication facet when it is governed by a communicative verb (\enquote{communicate}, \enquote{present}, \enquote{explain}, \enquote{report to}, \enquote{maintain relationships with}). Second, mentorship, which Section~\ref{sec:skill-taxonomy} admits as communication evidence when it is articulated as knowledge transfer, is recorded as a facet of its own so that its weight in the corpus becomes measurable. Third, leadership, which the annotation protocol treats as evidence for both skills, is recorded as the collaboration facet \enquote{team leadership}, with its communicative component captured by the mentoring facet.

\textbf{Procedure.} Each posting is segmented into clauses by splitting on line breaks, sentence punctuation, and bullet glyphs. A facet is assigned when a clause matches one of its patterns and no negative criterion matches within the same clause. The negative criteria encode the exclusions of Section~\ref{sec:skill-taxonomy}, namely machine-to-machine and system-level usages of the two surface forms (\enquote{communication protocols}, \enquote{GPU-to-GPU communication}, \enquote{collaborative filtering}, \enquote{collaboration tools} as a product feature) and descriptions of the workplace or the benefits package rather than of a required competency (\enquote{join our team}, \enquote{we are a small team}, \enquote{team events}, mentorship offered to the candidate, descriptions of the interview process). For every assignment, the matching clause is retained, so that each facet code carries its own evidence, and the coding can be audited clause by clause.

We implement this pass as a deterministic rule set rather than a second LLM application for three reasons. The detection decision, which is where contextual reasoning is indispensable, has already been made by the hybrid pipeline; attribution within an already-positive posting is a comparatively constrained task. A rule set is exactly reproducible and inspectable, which matters because the facet shares are descriptive quantities without a human-annotated ground truth of their own. And the rule set can be deliberately tuned for precision, so that reported facet shares are conservative.

\textbf{Derived measures.} Facets are not mutually exclusive, and a posting may express a requirement through several of them. We therefore additionally record \textit{breadth}, the number of distinct facets a posting expresses per skill, and we report the \textit{unattributed} residual, the positively labeled postings for which no facet fires. Finally, we code two specificity attributes over the communication-bearing clauses of each posting. Whether the counterpart of the communication is named (\textit{audience specified}: non-technical audiences, executives, clients, customers, end users, business stakeholders, regulators, peers, or juniors) and whether a channel is named (\textit{written}, \textit{verbal}, both, or neither).

\textbf{Validation.} We audited the coder by drawing a random sample of 120 evidence clauses, ten per facet, pooled across the ML and SE corpora, and reviewing each against the coding rules. Seven were judged misattributions, corresponding to a precision of approximately 94\%. The errors were concentrated in employer-branding passages that name a competency without requiring it (for example, \enquote{our business is built on connection, teamwork, and passion}), the residual of the negative criterion for environment descriptions. Because the coder is tuned for precision, the reported facet shares should be read as lower bounds; the unattributed residual, reported in Table~\ref{tab:facet-composition}, bounds the extent of under-attribution from above. The complete pattern set and the per-posting facet codes with their evidence clauses are included in the replication package.

\subsection{Statistical Analysis}

\begin{figure}
\centering
\includegraphics[width=1\textwidth]{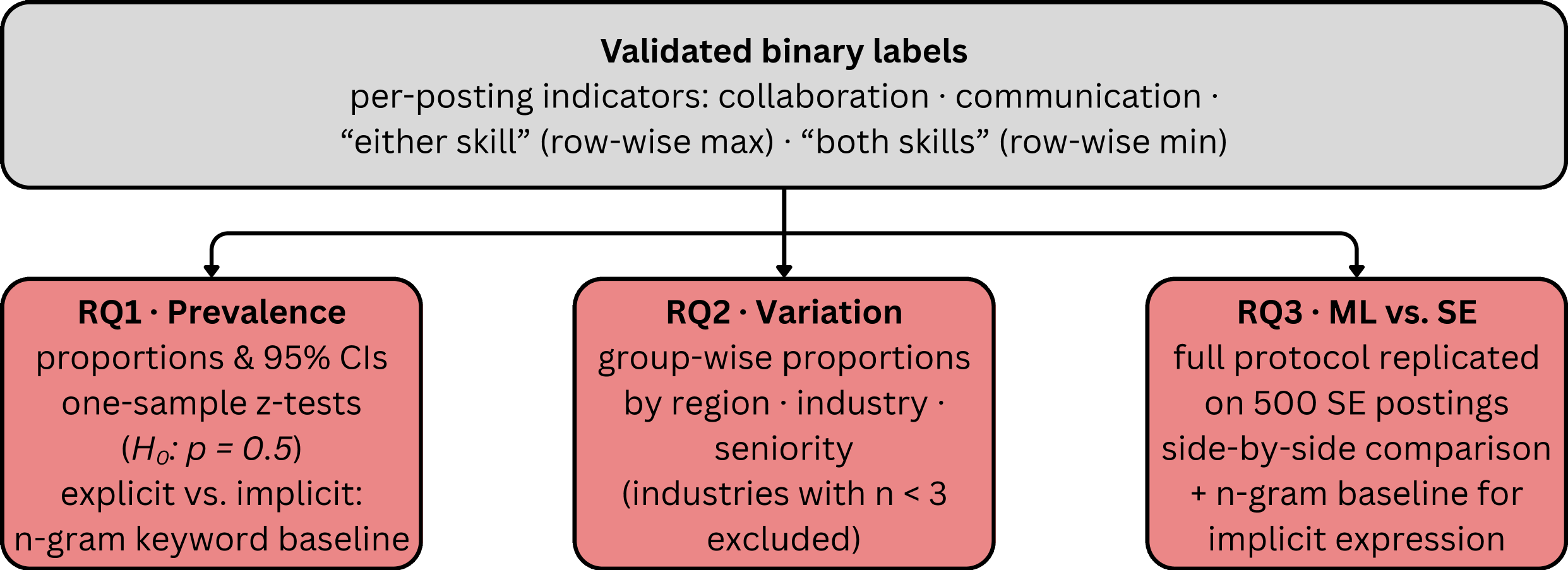}
\caption{An overview of the statistical analysis stage}
\label{stage4}
\end{figure}

The annotated dataset was loaded from an Excel workbook, and the collaboration and communication labels were encoded as binary indicators to ensure consistency in subsequent computations. Two derived variables were introduced to support a more nuanced view of skill demand: \enquote{either skill}, computed as the row-wise maximum of the two binary labels and indicating whether at least one of the target skills is required; and \enquote{both skills}, computed as the row-wise minimum and indicating whether both skills are simultaneously required. These derived indicators distinguish partial from combined skill requirements. Fig.~\ref{stage4} shows an overview of the statistical analysis stage. All analyses operate on the final hybrid human-LLM annotations, which serve as the ground truth.

Alongside the binary skill indicators, the facet codes of Section~\ref{sec:facet-coding} enter the analysis as twelve additional binary variables. Facet-level associations with the contextual dimensions of RQ2 and with the domain contrast of RQ3 are tested with the same procedure used for the skill indicators, that is $\chi^2$ tests of independence with Holm--Bonferroni adjustment and Cram\'er's $V$ as the effect size, with Fisher's exact test or Monte-Carlo resampling under fixed margins ($2\times 10^{5}$ resamples) wherever an expected cell count falls below five. Adjustment is applied within each family of twelve facet tests. Breadth, an ordinal count, is compared with Kruskal--Wallis tests across groups and with Mann--Whitney $U$ tests across domains.

\textbf{RQ1:} Overall prevalence of collaboration and communication in ML postings.

An analysis was conducted to estimate the proportions of postings mentioning collaboration, communication, both skills, and at least one of the two. Proportions were computed by dividing the number of positive cases by the total observations, and absolute frequencies were tabulated for mutually exclusive categories. To complement the analysis, a one-sample proportion z-test was applied to assess whether the proportion of postings mentioning at least one soft skill significantly exceeded a predefined benchmark, with separate tests and 95\% confidence intervals, computed as $\hat{p} \pm 1.96 \sqrt{\hat{p}(1-\hat{p})/n}$, for collaboration and communication individually. The one-sample proportion $z$-test is appropriate here because each indicator is a binary outcome observed over $n = 500$ independent postings, comfortably satisfying the normal approximation to the binomial, and the 50\% benchmark operationalizes the null position that a skill is no more likely to be required than not; rejection therefore establishes the skill as a majority requirement rather than an incidental one.

Throughout, descriptive statistics are complemented by one-sample proportion z-tests against a $P_0 = 0.5$ benchmark with a one-sided alternative $H_A : P > 0.5$, and by 95\% confidence intervals computed as $\hat{p} \pm 1.96 \sqrt{\hat{p}(1-\hat{p})/n}$. The hybrid human–LLM annotations defined before serve as the ground truth.

Additionally, to evaluate how effectively the human–LLM pipeline captures implicit expressions of soft skills, an n-gram baseline was constructed to represent only explicit mentions. The explicit baseline applies the regular-expression patterns shown in Listing~\ref{lst2}, covering surface forms such as \enquote{communication skills,} \enquote{verbal communication,} \enquote{team player,} and \enquote{cross-functional teams.} The hybrid annotation, by contrast, can identify implicit expressions in which the same competencies are signaled indirectly through phrases such as \enquote{explain solutions to stakeholders,} \enquote{liaise with partners internally and externally,} or \enquote{maintain successful customer relationships.} Comparing the two yields a measure of the contextual gain provided by the hybrid pipeline.

\begin{lstlisting}[caption={Regular-expression patterns used as the explicit-mention baseline.
},label={lst2}]
collaboration_patterns = [
  r"\bteam player\b",
  r"\bteamwork skills\b",
  r"\bcollaboration skills\b",
  r"\bcross-functional teams\b",
  r"\bcollaborative environment\b",
  r"\bwork with teams\b",
  r"\bworking with teams\b",
  r"\bcollaborate with\b",
  r"\bteam-oriented\b",
  r"\bteam environment\b",
  r"\bpartner with teams\b",
  r"\bworking collaboratively\b",
  r"\bcollaborative\b",
  r"\bcollaborate\b",
  r"\bcollaboration\b",
]
communication_patterns = [
  r"\bcommunication skills\b",
  r"\bstrong communication\b",
  r"\bexcellent communication\b",
  r"\bverbal communication\b",
  r"\bwritten communication\b",
  r"\binterpersonal skills\b",
  r"\binterpersonal communication\b",
  r"\bcommunicate effectively\b",
  r"\bgood communicator\b",
  r"\bpresentation skills\b",
  r"\bstakeholder communication\b",
  r"\bcommunicating with\b",
  r"\bcommunicate\b",
  r"\bcommunication\b",
]
\end{lstlisting}

\textbf{RQ2:} Variation across regions, industries, and seniority levels.

The dataset was partitioned along each contextual dimension, and group-level proportions
were computed via group-by aggregations on the binary skill variables. The within-group
means correspond to the proportion of postings requiring each skill and were converted
to percentages for cross-group comparability. The analysis was conducted separately for
collaboration, communication, and the \enquote{either skill} indicator, complemented by
absolute counts for each group. To complement the descriptive comparisons, each contextual dimension was tested for association with each skill indicator using chi-squared tests of independence on the corresponding $2 \times k$ contingency tables (region: $k = 5$; industry: $k = 4$; seniority: $k = 4$). Because this constitutes a family of nine related tests, p-values were adjusted with the Holm--Bonferroni procedure~\citep{Holm_1979}, which controls the family-wise error rate without assuming independence among tests; effect sizes are reported as Cram\'er's $V$. Where minimum expected cell counts fell below five (four of the nine tables), the chi-squared results were verified with Monte Carlo permutation tests ($10^{5}$ resamples), which yielded materially identical p-values. Given the quota-based sampling design described in Section~\ref{sec:data-collection}, these tests assess associations within the collected corpus and should not be read as licensing inference to the global population of ML postings.

\textbf{RQ3:} Cross-domain comparison of skill demand between ML and software engineering postings.

To determine whether the interpersonal demand observed for ML differs from that in the more established reference profession, the descriptive and inferential procedure used for RQ1 was replicated on the SE subset. After binary coding, proportions and absolute frequencies summarized the prevalence of collaboration and communication, as well as the \enquote{either skill} indicator. One-sample proportion z-tests were conducted under a null hypothesis $H_0 : p = 0.5$ with a one-sided alternative $H_1 : p > 0.5$, and 95\% confidence intervals were computed for each proportion. The same n-gram baseline used for RQ1 was applied to the SE subset to characterize the explicit-implicit gap and to enable direct comparison with the ML findings.

\section{Results}
\label{sec:results}

This section reports the empirical findings for the three research questions: the prevalence of collaboration and communication skills in ML job postings (RQ1), the variation of this demand across regions, industries, and seniority levels (RQ2), and the comparison of the ML findings with the SE subset (RQ3).

\subsection{Prevalence of Collaboration and Communication in ML Job Postings}

\begin{figure}
\includegraphics[width=1\textwidth]{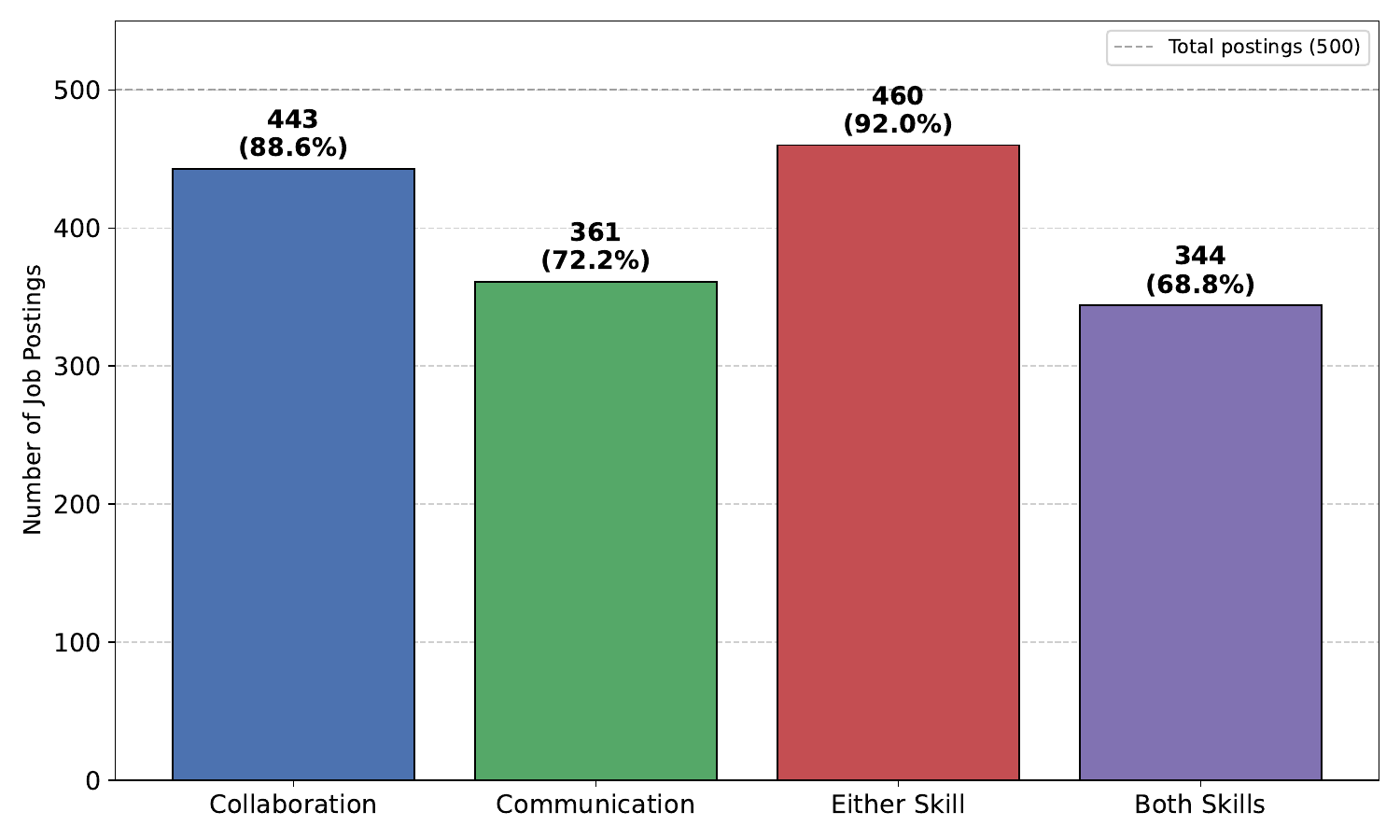}
\caption{Frequency of collaboration and communication skills in ML job postings}
\label{frequency-coco-in-ML-jobs}
\end{figure}

To answer \textbf{RQ1}, across the 500 ML postings, collaboration skills are present in 443 cases (88.6\%) and communication skills in 361 cases (72.2\%). At least one of the two skills appears in 460 job postings (92.0\%), and both skills are required simultaneously in 344 postings (68.8\%), as shown in Fig.~\ref{frequency-coco-in-ML-jobs}. Disaggregating into mutually exclusive categories, 99 mention collaboration alone, 17 ML job postings mention communication alone, 344 mention both, and only 40 mention neither. Collaboration is therefore the most frequently required individual competency, exceeding communication by 16.4\% points.

A one-sample proportion $z$-test on the \enquote{either skill} indicator strongly rejects the 50\% benchmark: $\hat{p} = 0.92$, $z = 18.78$, $p < 10^{-78}$, 95\% confidence interval $(0.896,\,0.944)$. The component skills are likewise required in significantly more than half of all postings:
\begin{itemize}
  \item Collaboration: $\hat{p} = 0.886$, $z = 17.26$, $p < 10^{-67}$,
        95\% confidence interval $(0.858,\,0.914)$.
  \item Communication: $\hat{p} = 0.722$, $z = 9.93$, $p < 10^{-22}$,
        95\% confidence interval $(0.683,\,0.761)$.
\end{itemize}
All tests reject $H_0$ at the $\alpha = 0.05$ level by a wide margin.

\textbf{Explicit vs.\ implicit expression.} To quantify how much of the observed demand is conveyed indirectly rather than through standard keywords, we compared the ground-truth labels against the n-gram baseline. Table~\ref{tab4} reports the resulting classification metrics. For both skills, the baseline achieves high precision ($\approx 0.99$) but materially lower recall, reflecting its tendency to miss context-dependent expressions while rarely producing false positives. For communication, 99 of the 361 ground-truth-positive postings (27.4\%) carry no explicit keyword and are recoverable only through the hybrid pipeline. For collaboration, 96 of the 443 ground-truth-positive postings (21.7\%) are similarly implicit. The corresponding Cohen's $\kappa$ values (0.410 for collaboration and 0.581 for communication) place the baseline only in the moderate range relative to the ground truth, confirming that explicit-keyword detection alone is insufficient for accurately characterizing collaboration and communication demand in ML job postings.

\begin{table}[t]
\centering
\caption{N-gram baseline performance against the hybrid human-LLM ground truth on the ML subset.}
\label{tab4}
\begin{tabular}{lrrrrrr}
\toprule
Skill & {Accuracy} & {Precision} & {Recall} & {F1} & {Cohen's $\kappa$} & {Implicit-only} \\
\midrule
Collaboration & 0.798 & 0.986 & 0.783 & 0.873 & 0.410 & 96 / 443 \\
Communication & 0.796 & 0.989 & 0.726 & 0.837 & 0.581 & 99 / 361 \\
\bottomrule
\end{tabular}
\end{table}

\subsubsection{Facet Composition of the Requirements}
\label{sec:rq1-facets}

Table~\ref{tab:facet-composition} and Fig.~\ref{facet-composition} report which facets carry the demand established above. Percentages in the table are shares of all 500 postings per domain; in the text, we additionally give shares of the postings that require the skill in question, which is the more natural denominator for compositional statements.

Collaboration requirements are dominated by two facets. General collaboration appears in 355 of the 443 collaboration-positive ML postings (80.1\%) and cross-functional collaboration in 267 (60.3\%), whereas teamwork and team practices account for only 125 (28.2\%), stakeholder coordination for 110 (24.8\%), and team leadership for 60 (13.5\%). The ordering is informative. The demand that ML postings articulate is predominantly a demand for integration across role boundaries, with data engineers, software engineers, product managers, and domain experts named as counterparties, rather than a demand for membership in a single delivery team. Cross-functional collaboration is required more than twice as often as teamwork. The familiar figure of the \enquote{team player} is thus far from the most common collaboration requirement in ML hiring.

Communication requirements are more evenly distributed across facets. General communication is the largest at 216 of 361 communication-positive postings (59.8\%), followed by audience-facing interaction (165, 45.7\%), presentation and explanation (160, 44.3\%), and mentoring and knowledge transfer (153, 42.4\%). Two facets are conspicuously rare. Written communication, which includes producing documentation, technical reports, and specifications, is required in only 67 postings (18.6\%), a striking figure for a field in which model, data, and experiment documentation is a recognized precondition for reproducibility and handover. Conflict and negotiation, a component of communication in the classifications we build on~\citep{Pol_kov_2023}, is essentially absent from the corpus (11 postings, 3.0\%).

\textbf{Breadth and specificity.} Postings that state a requirement rarely elaborate on it. Collaboration-positive ML postings express on average 2.07 of the five collaboration facets ($SD = 1.08$) and communication-positive postings 2.25 of the seven communication facets ($SD = 1.32$); 20.5\% and 28.0\% respectively express exactly one. In 58 collaboration-positive postings (13.1\%) and 41 communication-positive postings (11.4\%), the requirement is expressed only through the corresponding general facet, that is, through a bare formula such as \enquote{strong communication skills} with no further qualification. The specificity coding sharpens this observation: only 43.5\% of communication-positive ML postings name whom the candidate is expected to communicate with, and 42.1\% specify neither a written nor a verbal channel. Employers, therefore, assert the requirement far more often than they specify it.

\textbf{Attribution coverage.} The coder attributes 407 of the 443 collaboration-positive postings (91.9\%) and 337 of the 361 communication-positive postings (93.4\%) to at least one facet. The remaining 36 and 24 postings carry requirements phrased so indirectly that no facet pattern fires, for example, a solutions architect expected to \enquote{bridge technical discussions between clients and delivery teams}. Given the precision-oriented design of the coder, these residuals bound the under-attribution of the facet shares reported above.

\begin{tcolorbox}[title=Summary for RQ1, colback=gray!5!white, colframe=black!50!white]
Collaboration and communication are not peripheral attributes in ML postings, but core requirements present in the overwhelming majority of ML job advertisements, and roughly one in four of these requirements would be missed by traditional keyword-based pipelines. The facet decomposition shows what the aggregate figures conceal. Collaboration is demanded predominantly as integration across role boundaries (cross-functional collaboration, 60.3\% of collaboration-positive postings) rather than as membership in a delivery team (teamwork, 28.2\%). Communication is demanded first as a generic attribute (59.8\%) and then, in descending order, as audience-facing interaction (45.7\%), presentation and explanation (44.3\%), and mentoring (42.4\%), while written communication (18.6\%) and conflict management (3.0\%) are marginal. Requirements are asserted more often than they are specified. Fewer than half of the communication-positive postings name the audience, and two in five name no channel at all.
\end{tcolorbox}

\subsection{Variation Across Region, Industry, and Seniority}

Regarding \textbf{RQ2}, we now disaggregate the ML findings along the three contextual dimensions. For each dimension, group-level proportions were computed as the mean of the binary skill indicators, with absolute counts reported in parallel. Figs.~\ref{frequency-coco-in-ML-jobs-region}, \ref{frequency-coco-in-ML-jobs-sector}, and~\ref{frequency-coco-in-ML-jobs-seniority} show the resulting shares of postings that require collaboration, communication, and either skill, broken down by region, industry, and seniority level, respectively. Each figure is discussed in the corresponding subsection below.

Table~\ref{tab:rq2-tests} reports the corresponding tests of independence. After Holm--Bonferroni adjustment, only the association between region and communication remains statistically significant ($\chi^2(4) = 17.78$, adjusted $p = 0.012$, Cram\'er's $V = 0.19$). This conclusion is unchanged under alternative family definitions (adjusting within each dimension, or excluding the derived either-skill indicator). The seniority differences shown in Fig.~\ref{frequency-coco-in-ML-jobs-seniority}, most visibly the lower communication share at mid level (61.8\%, against 73.5--83.3\% in the other tiers), reach nominal significance for both skills (communication: $\chi^2(3) = 8.66$, $p = 0.034$; collaboration: $\chi^2(3) = 9.33$, $p = 0.025$) but do not survive adjustment ($p = 0.239$ and $p = 0.201$), and none of the industry comparisons approaches significance even before adjustment. The subgroup patterns described below should be read with this in mind. The data give firm evidence that regions differ in how they state communication requirements. They give suggestive but not conclusive evidence of differences across seniority levels, and no evidence that industries differ more than chance would explain. This fits two features of the data. Demand is high in every subgroup, and several subgroups are small.

Finally, prompted by the differentiated role structure of the ML labor market, we examined the concrete roles advertised in the job postings. A rule-based classification of the 500 job titles, manually verified by the annotators, shows that ML/AI engineers dominate the sample (256 postings, 51.2\%), followed by data scientists (97, 19.4\%), management and leadership roles (54, 10.8\%), MLOps/ML-infrastructure engineers (33, 6.6\%), solutions architects (28, 5.6\%), research scientists (25, 5.0\%), and data engineers (5, 1.0\%); two postings could not be assigned to a coherent role.

\begin{table}[t]
\centering
\caption{Chi-squared tests of independence between each contextual dimension and each
skill indicator on the ML subset ($n = 500$ per dimension), with Holm--Bonferroni
adjustment across the nine tests. For the four tables with minimum expected cell counts
below five, Monte Carlo permutation tests ($10^{5}$ resamples) confirmed the reported
p-values.}
\label{tab:rq2-tests}
\begin{tabular}{llrrrrr}
\toprule
Dimension & Indicator & $\chi^2$ & df & $p$ & Holm-adj.\ $p$ & Cram\'er's $V$ \\
\midrule
\multirow{3}{*}{Region}
 & Communication & 17.78 & 4 & 0.001 & 0.012 & 0.189 \\
 & Collaboration & 10.02 & 4 & 0.040 & 0.241 & 0.142 \\
 & Either skill  &  7.07 & 4 & 0.132 & 0.662 & 0.119 \\
\midrule
\multirow{3}{*}{Industry}
 & Communication &  5.15 & 3 & 0.161 & 0.662 & 0.101 \\
 & Collaboration &  3.83 & 3 & 0.281 & 0.741 & 0.088 \\
 & Either skill  &  3.16 & 3 & 0.368 & 0.741 & 0.080 \\
\midrule
\multirow{3}{*}{Seniority}
 & Communication &  8.66 & 3 & 0.034 & 0.239 & 0.132 \\
 & Collaboration &  9.33 & 3 & 0.025 & 0.201 & 0.137 \\
 & Either skill  &  4.14 & 3 & 0.247 & 0.741 & 0.091 \\
\bottomrule
\end{tabular}
\end{table}

\subsubsection{Geographic Region}

\begin{figure}
\includegraphics[width=1\textwidth]{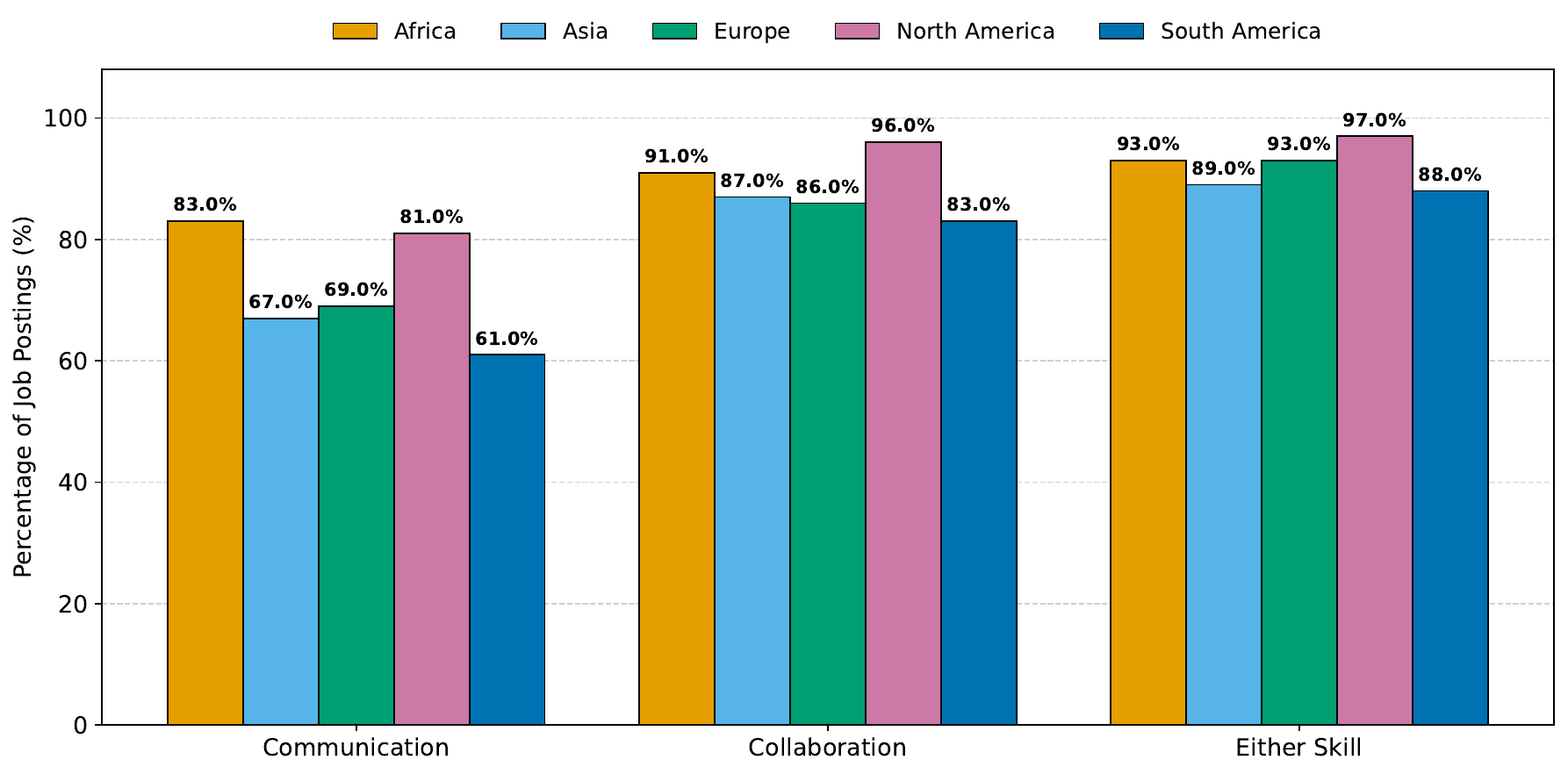}
\caption{Share of ML job postings requiring collaboration, communication, or at least one of the two (either skill), by region ($n = 100$ postings per region)}
\label{frequency-coco-in-ML-jobs-region}
\end{figure}

As illustrated in Fig.~\ref{frequency-coco-in-ML-jobs-region}, North America records the highest demand on collaboration (96.0\% of job postings) and on the \enquote{either skill} indicator (97.0\%), and the second-highest communication share (81.0\%), behind Africa (83.0\%). Europe and Africa form an intermediate tier, with the \enquote{either skill} indicator at 93.0\% in both regions but a noticeably lower communication share in Europe (69.0\%) than in Africa (83.0\%). Asia (89.0\% either) and South America (88.0\% either) exhibit the lowest explicit demand. The reduced prevalence in these regions may reflect a stronger emphasis on technical competencies in written advertisements, or cultural and organizational conventions that leave certain interpersonal expectations implicit rather than stated. The absolute-count visualization reinforces these patterns: Africa has the highest both-skill count (81), closely followed by North America (80), whereas South America has the lowest both-skill share (56 of 100 postings).

\textbf{Facet-level view.} Table~\ref{tab:facet-region} decomposes the regional pattern. The aggregate effect reported above, the only subgroup association to survive adjustment, resolves into a difference in which facets are named rather than a uniform difference in level. Audience-facing interaction (57\% in Africa and 42\% in North America against 21--23\% elsewhere; adjusted $p < 10^{-7}$, Cram\'er's $V = 0.30$) and presentation and explanation (57\% and 40\% against 19--23\%; adjusted $p < 10^{-8}$, $V = 0.31$) drive the aggregate result, joined by cross-functional collaboration (adjusted $p = 0.0004$, $V = 0.22$) and mentoring (adjusted $p = 0.001$, $V = 0.21$). Language proficiency runs in the opposite direction. It is stated in 20\% of European and 9\% of Asian and South American postings, but in only 4\% of North American and no African postings (adjusted $p < 0.0001$, $V = 0.24$). European postings, therefore, do not omit communication requirements so much as they state a different one, namely, the language in which the work will be conducted. Requirement breadth follows the same ordering, with African (2.33 collaboration facets, 2.83 communication facets) and North American postings (2.30, 2.49) elaborating requirements more than European (1.79, 2.00) or South American ones (1.77, 2.07); Kruskal--Wallis $p < 0.001$ for both skills. This reinforces the caution stated above: what varies across regions is the explicitness and the vocabulary of articulation, not demonstrably the underlying demand.

Overall, collaboration and communication skills are valued globally, but the explicit articulation of those expectations is markedly stronger in North America than elsewhere. This is an important caveat for cross-regional comparison. Lower visible demand does not necessarily indicate lower actual demand.

\subsubsection{Industry Domains}

\begin{figure}
\includegraphics[width=1\textwidth]{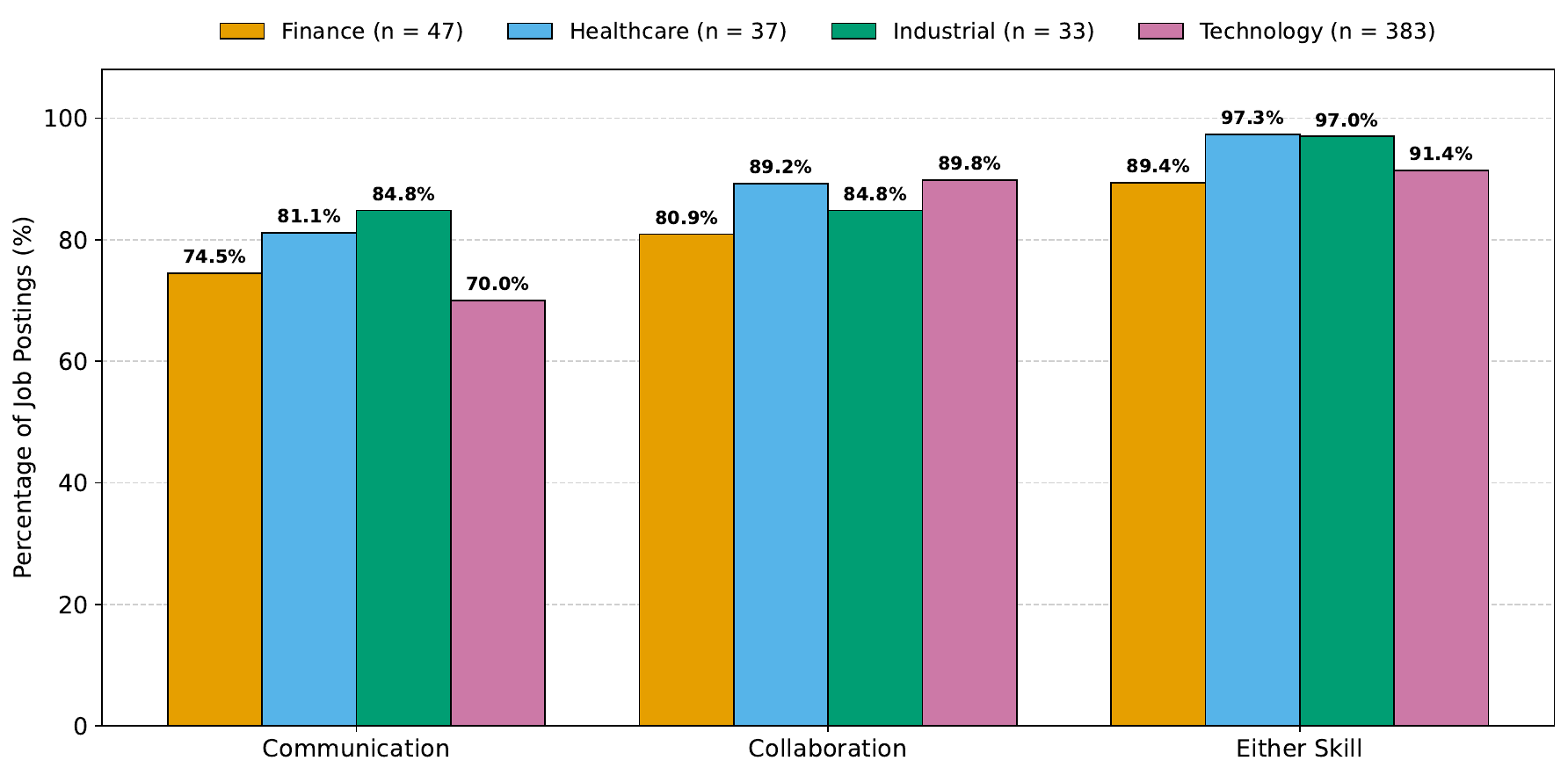}
\caption{Frequency of collaboration and communication skills in ML job postings by sector}
\label{frequency-coco-in-ML-jobs-sector}
\end{figure}

To avoid unreliable estimates from very small subgroups, the two smallest industry labels, manufacturing ($n = 22$) and energy ($n = 11$), were pooled into a single Industrial category. The resulting four categories, Finance ($n = 47$), Healthcare ($n = 37$), Industrial ($n = 33$), and Technology ($n = 383$), are reported in Fig.~\ref{frequency-coco-in-ML-jobs-sector}.

Healthcare exhibits the strongest demand on the \enquote{either skill} indicator (97.3\%) and the highest joint prevalence of both skills (73.0\%), consistent with the interdisciplinary and people-centered nature of medical work, where coordination across specialties and clear communication in high-stakes settings are constitutive of the role rather than ancillary. Industrial postings show essentially the same overall demand (97.0\%), a comparable joint prevalence of both skills (72.7\%), and the distinction of being the only sector in which the two skills are demanded at identical rates (84.8\% each), plausibly reflecting the coordination across engineering, operations, and production functions characteristic of industrial ML deployments and the cross-functional cooperation required to develop, deploy, and maintain ML systems in physical environments.

The Technology sector, while showing a slightly lower either-rate (91.4\%), accounts for the largest absolute number of postings requiring both skills (262 of 383). The volume of hiring in this sector, therefore, makes it the largest practical source of demand for combined collaboration and communication skills, even though the percentage prevalence is somewhat lower than in healthcare or industrial roles. Finance records the lowest demand on collaboration (80.9\%) and on the \enquote{either skill} indicator (89.4\%), with communication at 74.5\%; the lowest communication share is in fact recorded by the technology sector (70.0\%). All values nevertheless remain substantively high. The relative gap may reflect stronger emphasis on analytical and quantitative competencies in finance hiring, or a greater willingness to leave interpersonal expectations implicit.

\textbf{Facet-level view.} Although none of the industry comparisons at the level of the skill indicators approaches significance (Table~\ref{tab:rq2-tests}), one facet does vary systematically across sectors. Written communication is required in 36.4\% of industrial postings and 21.6\% of healthcare postings, against 11.2\% in technology and 8.5\% in finance (adjusted $p = 0.002$, Cram\'er's $V = 0.20$); no other facet survives adjustment within this family. The sectors in which ML systems are deployed in regulated, safety-relevant, or physically instrumented settings are thus the sectors that require candidates to produce documentation, specifications, and reports, consistent with the documentation obligations these settings impose. The sectoral signal in our corpus is therefore compositional. It concerns which interpersonal competency is named, not how often one is named at all.

The pattern across industries indicates that the intensity, rather than the presence, of collaboration and communication skills demand varies by sector. Healthcare and industrial roles, which require dense coordination and multidisciplinary interaction, articulate these expectations most strongly; technology operationalizes them at the largest scale; finance shows the weakest explicit emphasis while still requiring them in roughly 9 of 10 job postings.

\subsubsection{Seniority Level}

\begin{figure}
\includegraphics[width=1\textwidth]{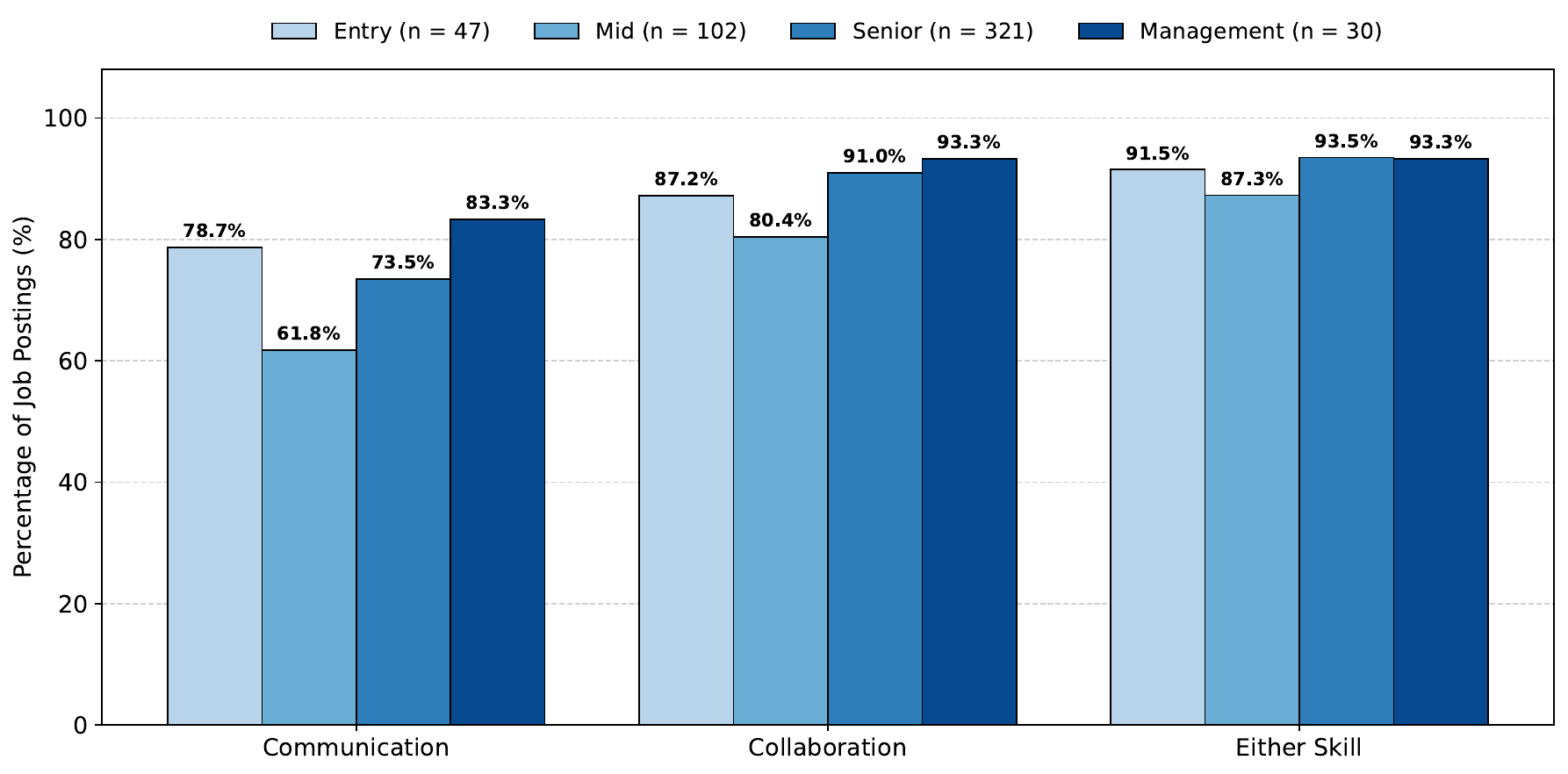}
\caption{Frequency of collaboration and communication skills in ML job postings by seniority level}
\label{frequency-coco-in-ML-jobs-seniority}
\end{figure}

Returning to Fig.~\ref{frequency-coco-in-ML-jobs-seniority}, the four categories follow a meaningful career-progression order: Entry, Mid, Senior, and Management.

Management roles record the highest individual proportions for both collaboration (93.3\%) and communication (83.3\%), aligning with the leadership emphasis on team coordination, stakeholder alignment, and decision facilitation. Senior roles record the highest \enquote{either skill} value (93.5\%), follow closely on collaboration (91.0\%), and contribute the largest absolute number of both-skill postings (228), consistent with the project-leadership, mentoring, and cross-team coordination expected at this level. Entry-level postings exhibit surprisingly strong demand, with 91.5\% requiring at least one skill, indicating that interpersonal competencies are treated as foundational from career start rather than as advanced attributes acquired through experience. The mid-level tier displays the lowest demand across all indicators, particularly for communication (61.8\%). One plausible interpretation is that mid-level postings concentrate explicit attention on technical execution and individual delivery while assuming that baseline interpersonal competencies have already been established by candidates at this stage; this would explain the dip in visible communication requirements without implying a decline in actual demand.

\textbf{Facet-level view.} Table~\ref{tab:facet-seniority} and Fig.~\ref{facet-by-seniority} show that seniority reorganizes the composition of the requirement rather than its level. Four facets vary significantly after adjustment, and three of them are the facets associated with responsibility for others. Team leadership rises from 0.0\% at entry level and 1.0\% at mid level to 14.3\% at senior level and 43.3\% in management postings (adjusted $p < 10^{-8}$, Cram\'er's $V = 0.31$); mentoring and knowledge transfer follows the same shape (17.0\%, 11.8\%, 36.4\%, 53.3\%; adjusted $p < 10^{-5}$, $V = 0.26$), as does stakeholder coordination (14.9\%, 12.7\%, 23.7\%, 46.7\%; adjusted $p = 0.005$, $V = 0.19$). The fourth, general collaboration, varies more weakly and non-monotonically (66.0\%, 58.8\%, 76.3\%, 63.3\%; adjusted $p = 0.038$, $V = 0.16$), and is not a responsibility-for-others facet. General communication, by contrast, does not vary significantly across the four tiers (48.9\%, 33.3\%, 45.8\%, 40.0\%), and neither does audience-facing interaction or presentation and explanation. Entry-level postings are, descriptively, the group with the highest prevalence of teamwork (36.2\%) and of written communication (19.1\%), though neither association survives adjustment.

Two consequences follow for the interpretation of the U-shaped aggregate profile. First, the intensification at the senior and management levels is almost entirely accounted for by facets that presuppose responsibility for other people, not by a general increase in interpersonal demand. Second, the mid-level dip is not accompanied by a comparable narrowing of what is asked. Among postings that state a collaboration requirement, mid-level postings name somewhat fewer facets (1.70) than entry-level (1.90), senior (2.15), or management postings (2.54), Kruskal--Wallis $p = 0.001$, whereas for communication the breadth differences are not significant ($p = 0.128$). The mid-level dip is therefore best read as the near-absence of the responsibility-for-others facets in that tier combined with a slightly lower propensity to state a requirement at all, rather than as a distinct mid-career communication regime.

Taken together, the seniority data describe a non-monotone but interpretable profile. High demand at career entry (foundational expectation), a relative dip at mid-level (technical focus), and renewed intensification at senior and management levels (leadership and coordination demands). The pattern indicates that the role of collaboration and communication evolves with career progression, becoming most prominent in positions with broader coordination and leadership responsibilities.

\subsubsection{Advertised Role}
\label{sec:rq2-role}

The role classification introduced above affords a fourth contextual dimension, and it produces the largest effects observed anywhere in this study. Fig.~\ref{facet-by-role} reports the facet profile of the six role groups with at least twenty postings. Six facets vary significantly across roles after adjustment. Team leadership (adjusted $p < 10^{-12}$, Cram\'er's $V = 0.38$), stakeholder coordination ($p < 10^{-9}$, $V = 0.35$), presentation and explanation ($p < 10^{-7}$, $V = 0.32$), audience-facing interaction ($p < 10^{-7}$, $V = 0.31$), general communication ($p = 0.009$, $V = 0.20$), and mentoring ($p = 0.014$, $V = 0.20$).

The profiles are ordered along an outward-facing axis. Solutions architects sit at one extreme, with stakeholder coordination in 67.9\%, audience-facing interaction in 75.0\%, and presentation and explanation in 71.4\% of postings, and they elaborate their requirements most extensively (2.67 collaboration and 3.20 facets per posting requiring the respective skill). MLOps and ML-infrastructure engineers sit at the other end, with each of these three facets in 9.1\% of postings and general communication in 18.2\%. Data scientists occupy an intermediate position but lean towards explanation (46.4\%) and audience-facing interaction (41.2\%) while recording the lowest teamwork share of any role (12.4\%); ML and AI engineers, the largest group, are close to the corpus average on the intra-team collaboration facets (cross-functional collaboration 53.9\%, teamwork 28.9\%) but below it on every outward-facing facet; management postings are distinguished chiefly by team leadership (46.3\%) and mentoring (42.6\%). Summarizing each posting by whether it expresses any outward-facing facet, that is, stakeholder coordination, audience-facing interaction, or presentation and explanation, yields 85.7\% for solutions architects, 68.5\% for management, 64.0\% for research scientists, 56.7\% for data scientists, 40.6\% for ML and AI engineers, and 12.1\% for MLOps engineers.

Two caveats apply. Role and seniority are not independent, since management postings are by construction concentrated in the management tier, and the associations reported here are marginal rather than adjusted for one another. The two smallest groups, research scientists ($n = 25$) and solutions architects ($n = 28$), also have correspondingly wide confidence intervals. The ordering is nevertheless robust to these caveats in the sense that the effect sizes for role exceed those of every other dimension we examine, including the ML--SE contrast reported in the next section.

\begin{tcolorbox}[title=Summary for RQ2, colback=gray!5!white, colframe=black!50!white]
Across region, industry, and seniority, the \enquote{either skill} prevalence remains above 87\% in every subgroup, and at the level of the skill indicators, only the regional variation in communication survives adjustment for multiple testing. The facet decomposition shows that the contextual dimensions nevertheless act systematically, by reorganizing the composition of the requirement. Regionally, African and North American postings name audience-facing and presentation facets two to three times as often as other regions, whereas European postings substitute language proficiency. Across seniority, the four facets that vary significantly are exactly those that presuppose responsibility for others (leadership, mentoring, stakeholder coordination), while general communication is flat. Across industries, where no aggregate association was detectable, written communication is significantly more prevalent in industrial and healthcare postings. The advertised role produces the largest effects in the study, separating outward-facing solutions architects from inward-facing MLOps engineers by a wide margin.
\end{tcolorbox}

\subsection{Comparison with SE Job Postings}
\label{sec:comparison-with-SE-jobs}

\begin{figure}
\includegraphics[width=1\textwidth]{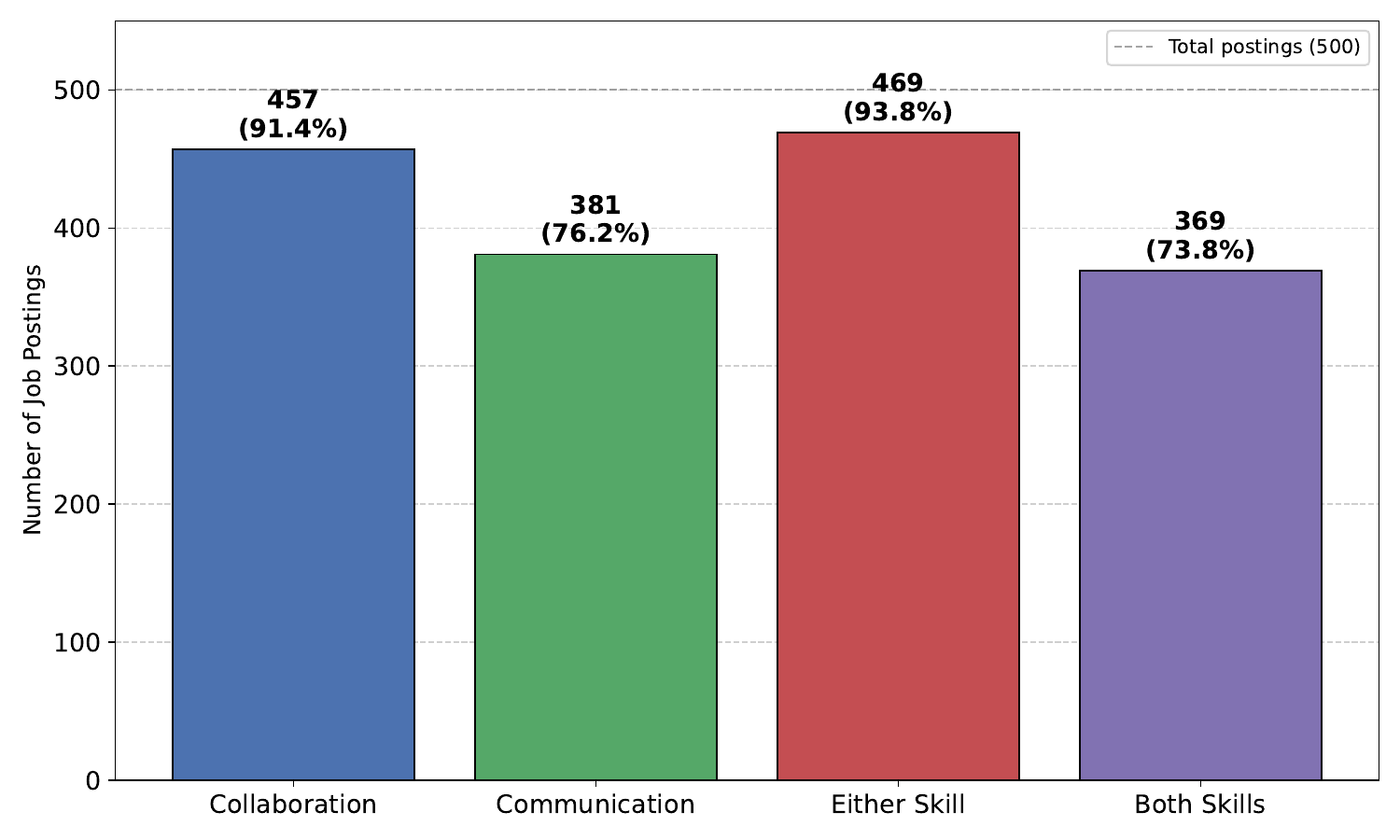}
\caption{Frequency of collaboration and communication skills in SE job postings}
\label{frequency-coco-in-SE-jobs}
\end{figure}

To assess whether ML and SE roles differ in their collaboration and communication requirements and answer \textbf{RQ3}, the labeling procedure was replicated on 500 SE postings. The resulting prevalences are visualized in Fig.~\ref{frequency-coco-in-SE-jobs}.

Collaboration skills appear in 457 SE job postings (91.4\%) and communication skills in 381 (76.2\%). At least one of the two skills is required in 469 postings (93.8\%), and both are required simultaneously in 369 (73.8\%). The mutually exclusive distribution comprises 12 postings with communication only, 88 with collaboration only, 369 with both, and 31 with neither.

As in RQ1, each proportion was tested against a 50\% benchmark, which operationalizes the null position that the skill is no more likely to be required than not; rejecting this null establishes the skill as a majority requirement of SE hiring rather than an incidental attribute. All three proportions significantly exceed the benchmark:

\begin{itemize}
  \item Either skill: $\hat{p} = 0.938$, $z = 19.59$, $p < 10^{-86}$, 95\% confidence interval $(0.917,\,0.959)$.
  \item Collaboration: $\hat{p} = 0.914$, $z = 18.51$, $p < 10^{-77}$, 95\% confidence interval $(0.889,\,0.939)$.
  \item Communication: $\hat{p} = 0.762$, $z = 11.72$, $p < 10^{-32}$, 95\% confidence interval $(0.725,\,0.799)$.
\end{itemize}

In substantive terms, even the lower bounds of the confidence intervals imply that at least 91.7\% of SE postings in our corpus require one of the two skills, at least 88.9\% require collaboration, and at least 72.5\% require communication; the extreme $z$-statistics indicate that these majorities cannot be attributed to sampling variation within the corpus. SE thus reproduces the qualitative pattern established for ML in RQ1: both competencies are majority requirements, and collaboration is the more consistently demanded of the two. This equivalence in statistical footing is what licenses the direct cross-domain comparison that follows.

\textbf{Cross-domain comparison.} Table~\ref{tab5} places the ML and SE findings side by side.

\begin{table}[t]
  \centering
  \caption{Side-by-side comparison of collaboration and communication prevalence in ML and SE postings.}
  \label{tab5}
  \begin{tabular}{lrrr}
    \toprule
    Indicator     & {ML (\%)} & {SE (\%)} & {Difference (pp)} \\
    \midrule
    Collaboration & 88.6 & 91.4 & +2.8 \\
    Communication & 72.2 & 76.2 & +4.0 \\
    Either skill  & 92.0 & 93.8 & +1.8 \\
    Both skills   & 68.8 & 73.8 & +5.0 \\
    \bottomrule
  \end{tabular}
\end{table}

SE job postings show marginally higher prevalence on every indicator, but the gaps are small; the largest, 5.0\% points for both skills, is well within the overlap of the corresponding 95\% confidence intervals (ML: $(0.647,\,0.729)$; SE: $(0.700,\,0.776)$). In both domains, collaboration is emphasized more strongly than communication, reinforcing the interpretation that team-oriented competencies are central to technical roles regardless of subfield. The slightly higher SE values may reflect the iterative, team-based nature of software development, where coordination across developers, testers, designers, and stakeholders is constitutive of the workflow. The near parity with ML, where similar interdisciplinary collaboration is required across data scientists, ML engineers, product managers, and domain specialists, suggests that the data-driven character of ML does not impose qualitatively distinct interpersonal demands; rather, both domains converge on a high baseline of collaboration and communication skills expectations.

\textbf{Facet composition across domains.} The near parity of the aggregate prevalences conceals a systematic difference in how the two domains express their requirements. Table~\ref{tab:facet-composition} and Fig.~\ref{facet-composition} report the facet shares side by side. Four contrasts survive Holm--Bonferroni adjustment across the twelve facets. SE postings articulate collaboration as teamwork and team practices far more often than ML postings (41.2\% against 25.0\% of all postings; adjusted $p < 10^{-6}$, Cram\'er's $V = 0.17$) and state language proficiency about twice as often (17.6\% against 8.4\%; adjusted $p = 0.0002$, $V = 0.14$). ML postings, conversely, ask more often for presentation and explanation (32.0\% against 18.6\%; adjusted $p < 10^{-4}$, $V = 0.15$) and for audience-facing interaction (33.0\% against 23.6\%; adjusted $p = 0.009$, $V = 0.10$). Two further contrasts point in the same direction without reaching adjusted significance. Cross-functional collaboration is more prevalent in ML (53.4\% against 45.6\%, adjusted $p = 0.095$) and mentoring in SE (38.2\% against 30.6\%, adjusted $p = 0.091$).

This divergence is not a matter of one domain simply saying more. Postings that state a requirement elaborate it to the same extent in both domains, with 2.07 against 2.04 collaboration facets (Mann--Whitney $p = 0.52$) and 2.25 against 2.07 communication facets ($p = 0.09$). Summarizing each posting by whether it expresses any outward-facing facet yields 49.0\% for ML against 41.2\% for SE ($\chi^2(1) = 6.14$, $p = 0.013$, $V = 0.08$), and the audience of the communication is named in 43.5\% of communication-positive ML postings against 29.9\% of SE ones ($\chi^2(1) = 14.72$, $p = 0.0001$). The two domains, therefore, demand interpersonal competence at the same level and with the same degree of elaboration, but they orient it differently. SE postings describe an inward demand, directed at the delivery team and its practices, while ML postings describe an outward one, directed at the parties for whom analytical work has to be made intelligible.

\textbf{Explicit vs. implicit expression in SE.} Applying the same n-gram baseline used for RQ1 to the SE subset (Table~\ref{tab6}) yields a result parallel to that observed for ML postings. High precision, materially lower recall, and a substantial number of ground-truth-positive postings that are not detectable through keyword matching alone. Specifically, 132 of the 457 collaboration-positive SE postings (28.9\%) and 123 of the 381 communication-positive postings (32.3\%) express the requirement implicitly. Both implicit shares are slightly higher than the corresponding ML values (21.7\% and 27.4\%), suggesting that SE job advertisements may rely on indirect language at least as much as ML postings, if not more. The very low Cohen's $\kappa$ for the collaboration baseline (0.289) particularly highlights the limitations of keyword detection for this construct on the SE subset.

Table~\ref{tab:facet-implicit} shows further that implicit expression is not distributed at random across facets. Among the ML postings whose communication requirement is invisible to the n-gram baseline, mentoring and knowledge transfer is the most frequent facet (51.5\%, against 38.9\% among the explicitly detected postings), followed by presentation and explanation (29.3\%) and audience-facing interaction (24.2\%), while general communication is nearly absent (4.0\%), which is expected given that the baseline patterns target precisely that facet. Language proficiency is likewise over-represented in the implicit set (19.2\% against 8.8\%). The same ordering holds in the SE corpus. Keyword pipelines, in other words, do not lose a random quarter of the requirements; they systematically lose the requirements that are expressed as duties, such as mentoring colleagues or explaining results to stakeholders, and those expressed as language requirements, and they retain the requirements expressed as attributes of the candidate.

\begin{table}[t]
\centering
\caption{N-gram baseline performance against the hybrid human--LLM ground truth on the SE subset.}
\label{tab6}
\begin{tabular}{lrrrrrr}
\toprule
Skill & {Accuracy} & {Precision} & {Recall} & {F1} & {Cohen's $\kappa$} & {Implicit-only} \\
\midrule
Collaboration & 0.734 & 0.997 & 0.711 & 0.830 & 0.289 & 132 / 457 \\
Communication & 0.728 & 0.952 & 0.677 & 0.791 & 0.431 & 123 / 381 \\
\bottomrule
\end{tabular}
\end{table}

\begin{tcolorbox}[title=Summary for RQ3, colback=gray!5!white, colframe=black!50!white]
Collaboration and communication are required at similarly high levels in ML and SE postings, and the marginal advantage of SE, on the order of 2--5 percentage points, does not constitute a substantive divergence in employer expectations. The facet decomposition qualifies this equivalence in an important way. At equal levels and equal breadth of articulation, the two domains orient their interpersonal demand differently. SE postings emphasize teamwork and team practices, whereas ML postings emphasize presentation, explanation, and audience-facing interaction, and name the audience of the communication about 1.5 times as often as SE postings (43.5\% vs. 29.9\%). ML and SE postings thus require these skills equally often, but they ask for different kinds of collaboration and communication.
\end{tcolorbox}

\section{Discussion}
\label{sec:discussion}

This section interprets the empirical findings against the prior literature, articulates their implications for the four stakeholder groups, reflects on the methodological lessons emerging from the hybrid annotation pipeline, and addresses several interpretive questions that the data invite.

\subsection{Soft Skills as Foundational Rather than Peripheral Requirements}

The first and most consistent finding of this study is that collaboration and communication skills are not peripheral attributes of ML hiring but core requirements. Across the ML corpus, at least one of the two skills was demanded in 92.0\% of job postings. This result strongly confirms a pattern specifically for ML that the broader labor-market literature has documented across other domains;~\citet{Deming_2018} reported a rising wage premium for occupations combining cognitive and social skills,~\citet{Lyu_2021} observed parallel growth in soft-skill demand in the U.S. energy sector, and~\citet{Pol_kov_2023} estimated that soft skills account for more than 40\% of competency requirements in Slovak technical job postings. Our findings indicate that ML, often perceived as a quintessentially technical and individualistic profession, follows the same trajectory.

What we add to this consensus is empirical specificity. Prior work on ML and AI roles, such as~\citet{Vu_2026}, has reported soft-skill cluster prevalences in the 19–21\% range, treating ML as a subset of a larger AI aggregate. Our targeted analysis suggests these aggregate-level figures substantially understate the ML-specific demand. Once the analysis is restricted to ML postings and the detection method captures implicit expression, the prevalence rises to 88.6\% for collaboration and 72.2\% for communication. This gap is not merely a methodological artifact; it reflects what aggregate analyses miss when they fold ML into broader categories and rely on surface-level keyword matching.

A second nuance concerns the relative weighting of the two skills. Collaboration is demanded in 88.6\% of ML postings against 72.2\% for communication, a 16.4\% point gap. This ordering is consistent with the data scientist study by~\citet{De_Morais_Le_a_2025, Azamnouri_2026_b}, which identified collaboration as the most consistently surfaced soft skill in their mixed-method analysis. We interpret the asymmetry as reflecting the interdisciplinary structure of ML work. Collaboration is required to integrate with data engineers, software developers, product managers, and domain specialists, while communication is required only when ML practitioners are expected to externalize their work to non-technical stakeholders. The two roles overlap but are not identical, and the data suggest that ML job postings emphasize the integrative role over the externalizing one.

\subsection{The Composition of Interpersonal Demand}
\label{sec:discussion-composition}

The prevalence findings confirm, for ML specifically, a pattern that the broader labor-market literature has established elsewhere. The facet decomposition is where our data depart from that literature, and three findings deserve emphasis.

\textbf{Integration, not membership.} Beyond generic statements of collaboration, the most common collaboration requirement in ML postings is not participation in a team but integration across role boundaries. Cross-functional collaboration is articulated in 60.3\% of collaboration-positive postings against 28.2\% for teamwork, and the counterparties named are overwhelmingly other functions: data engineers, software engineers, product managers, designers, and domain experts. This is a direct labor-market counterpart to the interface problems that qualitative studies of ML practice report as the dominant source of friction in production ML projects~\citep{Nahar_2022, Mailach_2023}. Where those studies observe coordination failures at the boundaries between data scientists, engineers, and domain experts, employers appear to have internalized the diagnosis and to be hiring against it. The practical implication differs from the one usually drawn from soft-skill research: what ML curricula and hiring processes need to develop is not generic teamwork but the ability to work across an expertise boundary.

\textbf{An outward orientation that distinguishes ML from SE at equal prevalence.} The comparison with SE is where the value of facet-level analysis is clearest. At the level of the skill indicators, the two domains are indistinguishable, and the natural conclusion, which we drew in the preceding section, is that interpersonal demand is a property of modern technical work rather than of ML in particular. The facet composition shows that this conclusion holds for the level of demand but not for its content. SE postings are oriented inward, towards the delivery team and its practices; ML postings are oriented outward, towards the parties to whom analytical work must be made intelligible. The effect sizes are modest ($V = 0.10$ to $0.17$), but they are consistent in direction across four facets, and they survive adjustment, and they are not an artifact of one domain elaborating its requirements more fully, since breadth is statistically indistinguishable between the two corpora. The substantive reading is that the interdisciplinary character of ML work, which the introduction posits, does register in hiring language, but as a reorientation of interpersonal demand rather than as an increase in it.

\textbf{A specificity gap.} Employers assert interpersonal requirements far more often than they specify them. In 11.4\% of communication-positive ML postings, the requirement is carried entirely by a generic formula, fewer than half name the audience, and two in five name no channel. Written communication is required in only 18.6\% of communication-positive postings, which sits uneasily with the documentation obligations that reproducible ML practice and model handover impose, and conflict management is effectively absent even though the classifications we build on treat it as a component of communication. The gap matters practically because a requirement that is asserted but not specified cannot be assessed. Neither the applicant preparing an application nor the screening process evaluating it can act on \enquote{strong communication skills} in the way it can act on \enquote{able to present model limitations to clinical staff}.

\subsection{Structure of Variation: Region, Industry, and Seniority}

While the baseline demand for soft skills is high across the corpus, the intensity of demand varies meaningfully along the three contextual dimensions examined in RQ2. The pattern is not best characterized as a uniform global signal but as a structured variation along three axes that interact in interpretable ways.

Regional variation reflects articulation conventions, not necessarily underlying demand. North America records the highest demand on collaboration and on the \enquote{either skill} indicator, Africa the highest on communication, while South America records the lowest on every indicator. This is the only subgroup effect that survives adjustment for multiple testing (communication: $\chi^2(4) = 17.78$, adjusted $p=0.012$; Table~\ref{tab:rq2-tests}). It would be tempting to conclude that North American ML workplaces are uniquely collaborative, but a more cautious interpretation is required. Prior work on cross-cultural HR practices~\citep{Bee_2015, Zainal_Shah_2022} suggests that the articulation of soft-skill requirements in job advertisements varies systematically with labor-market conventions, and that the gap between stated and actual demand can be substantial. Our results are consistent with this interpretation. Even the lowest-prevalence regions (South America, Asia) still record either-skill demand near 88\%, indicating that the underlying expectation is high everywhere; what differs is the explicitness with which it is stated. This caveat matters for both researchers and recruiters. A comparative claim, such as ML roles in North America require more interpersonal skill than ML roles in Asia, is not supported by these data; what is supported is that ML job postings in North America articulate that requirement more visibly.

Furthermore, industry variation reflects the inherent structure of work. The healthcare and industrial sectors record the strongest demand, the technology sector records the largest absolute volume, and finance records the lowest explicit demand. The healthcare and industrial patterns are consistent with the interpretation that interpersonal competencies are constitutive of the role itself. Medical ML work requires coordination across specialties in high-stakes environments, and industrial ML work requires cross-functional cooperation between engineering, operations, and production. The finance pattern is more interesting interpretively. Finance is one of two sectors in our corpus where the gap between collaboration and communication compresses, to 6.4 percentage points against 19.8 in technology; in industrial postings, the gap closes entirely, with both skills demanded in 84.8\% of postings. One plausible reading is that finance hiring relatively weights technical and quantitative competencies more heavily than interpersonal ones, leaving the two soft skills demanded at similar but lower levels. Another is that finance is sufficiently structured that explicit articulation of interpersonal expectations is less necessary. Disentangling these interpretations would require qualitative follow-up with finance recruiters. None of the industry differences is statistically distinguishable from chance (Table~\ref{tab:rq2-tests}), so the sectoral patterns described here are descriptive tendencies within our sample.

Moreover, seniority variation describes a non-monotone profile. Across the four seniority tiers, soft-skill demand follows a U-shape: high at entry, lowest at mid-level, then rising again at senior and management tiers. This profile is descriptive, the association reaches nominal significance for both skills but does not survive multiplicity adjustment (Table~\ref{tab:rq2-tests}), which is itself consistent with the interpretation below that the dip reflects advertisement language rather than a robust shift in demand. This profile is unusual relative to the linear narrative that soft skills become more important with seniority, which dominates the practitioner literature. Collaboration and communication skills are treated as foundational at career entry, recede in explicit visibility during the technical-execution phase of mid-level work, and then return prominently at senior and management levels where coordination and leadership responsibilities dominate. The mid-level dip in communication is the most striking feature. We suggest two interpretations: either employers assume that mid-level candidates have already established baseline communication ability through prior career stages and therefore omit it from explicit requirements, or mid-level postings are functionally narrower and concentrate on technical execution. Both interpretations imply that the mid-level dip is an artifact of advertisement language rather than a genuine decline in workplace demand.

The facet analysis also revises two of the interpretations offered above. The seniority profile is not a fluctuation in the amount of interpersonal demand across a career but a change in its object: the facets that increase with seniority are those that presuppose responsibility for other people (team leadership, mentoring, stakeholder coordination), while the facets describing communication as such remain flat across all four tiers. The mid-level dip is correspondingly better described as the tier in which responsibility-for-others facets are not yet named than as a tier in which communication matters less. The industry dimension, which the aggregate analysis found indistinguishable from chance, does carry a signal once facets are separated. Written communication is three times as prevalent in industrial postings, and twice as prevalent in healthcare postings, as in technology postings, which is consistent with documentation obligations in regulated and safety-relevant deployment settings and offers a more specific explanation than the notion of \enquote{high-stakes coordination} we proposed above.

Perhaps the most consequential structural finding is that the heterogeneity within ML exceeds the heterogeneity between ML and SE. The role dimension produces effect sizes of $V = 0.31$ to $0.38$ on the outward-facing facets and on leadership, against $V = 0.10$ to $0.17$ for the ML--SE contrast. An MLOps posting and a solutions-architect posting, both of which our sampling counts as ML roles, differ in their interpersonal profile far more than the ML and SE corpora differ from one another. Treating \enquote{ML roles} as a single occupational category, as the aggregate literature does, therefore averages over a distinction that is larger than the one that literature sets out to examine.

\subsection{Structural Differences Between ML and SE Job Postings}

Our analysis for RQ3 indicates that SE postings reference collaboration and communication skills marginally more often than ML postings. We caution, however, against reading this gap as direct evidence that SE employers attach greater importance to these two skills. A more plausible explanation lies in the maturity of the two occupational categories and in the linguistic conventions that follow from that maturity. SE is a comparatively more established profession for which hiring practices have produced a fairly standardized template for job advertisements; competencies tend to be enumerated explicitly, often under a dedicated heading, and recruiters frequently rely on reusable text blocks across postings. ML roles, by contrast, are comparatively recent on the labor market, and the corresponding advertisements vary substantially in how responsibilities, qualifications, and team expectations are framed.

This structural asymmetry suggests two possible consequences, only one of which our data can speak to directly. First, an explicit collaboration or communication bullet point may be more likely to appear in an SE posting simply because such postings are more formulaic, regardless of the importance the employer attaches to the underlying competency. We did not annotate structural features such as posting length or the presence of a dedicated requirements section, so this mechanism remains a hypothesis rather than a demonstrated effect. Second, one might expect implicit cues, phrases that describe interpersonal behavior without naming a skill, to be more prevalent in ML postings if their authors write in a more narrative register. Our data do not support this expectation. Comparing the implicit shares in Tables~\ref{tab4} and~\ref{tab6}, indirect expression is in fact more common in SE than in ML for both skills (collaboration: 28.9\% vs.\ 21.7\%; communication: 32.3\% vs.\ 27.4\%). SE postings thus reference the two competencies more often than ML postings in both the explicit and the implicit register, which rules out the possibility that the cross-domain gap is an artifact of ML requirements being hidden in indirect language.

Two conclusions follow. Substantively, because the prevalence gap is small and falls within the overlap of the corresponding confidence intervals (Section~\ref{sec:comparison-with-SE-jobs}), and because it persists across both registers of expression, the most defensible reading remains that the two domains converge on an equivalently high baseline of interpersonal demand rather than that SE employers value these skills more. Methodologically, since the maturity-driven mechanism proposed above was not directly measured, we recommend that future cross-domain comparisons annotate posting length and structural features, e.g., the presence of a dedicated requirements section, so that convention effects can be separated from demand effects, and continue to use extraction methods that explicitly model implicit cues, as we discuss next.

\subsection{Comparative Assessment of Classification Methods}
\label{sec:method-comparison}

As reviewed in Section~\ref{sec:methods-for-skill-extraction}, existing skill-extraction methods range from keyword and dictionary matching, which is computationally cheap and trivially scalable but blind to implicit expression, through embedding-based extractors, to neural sequence-labeling models such as SkillSpan~\citep{Zhang_2022}, which locate skill-mentioning spans with high precision but neither judge whether an entire posting implies a competency nor operate without task-specific labeled data. LLM-as-a-Judge reframes the task as contextual semantic evaluation. The model reads the complete posting and judges whether the role implies the target competency, so that a duty such as \enquote{explain technical solutions to business stakeholders} counts as evidence of communication ability even though the term \enquote{communication} never appears, and the framework can be retargeted to a new skill or domain by editing the prompt rather than re-labeling a corpus. To make the added value of this reframing empirically transparent, we benchmarked it against the two extremes most commonly used in practice: a lexical n-gram baseline and a fine-tuned BERT classifier.

\textbf{Comparison with the n-gram baseline.}
To quantify the practical benefit of contextual reasoning, we compared LLM-as-a-Judge against a lexical n-gram baseline that assigns collaboration and communication labels on the basis of curated keyword lists. Agreement with the two independent human annotators was measured using Cohen's $\kappa$ on both the ML and the SE subsets; Table~\ref{tab:method-comparison} consolidates the results. The baseline reaches only fair-to-moderate agreement ($\kappa \approx 0.28$--$0.58$ across domains and skills), whereas LLM-as-a-Judge is consistently substantial ($\kappa \approx 0.70$--$0.81$). Two observations are worth emphasizing. First, the gain is positive in all eight comparisons, across domains, annotators, and skill categories, which rules out a dataset-specific artifact. Second, the gap is largest for collaboration ($\Delta = +0.40$ on ML, $+0.43$ on SE), the category for which our qualitative inspection revealed the highest prevalence of implicit cues; this is precisely the pattern in which contextual reasoning is expected to outperform lexical matching.

\begin{table}[t]
\centering
\caption{Cohen's $\kappa$ agreement of the n-gram baseline and LLM-as-a-Judge with the two human annotators (A1, A2), by domain and skill. $\Delta$ denotes the mean improvement of LLM-as-a-Judge over the baseline across the two annotators.}
\label{tab:method-comparison}
\small
\begin{tabular}{llccccc}
\toprule
 & & \multicolumn{2}{c}{n-gram baseline} & \multicolumn{2}{c}{LLM-as-a-Judge} & \\
\cmidrule(lr){3-4}\cmidrule(lr){5-6}
Domain & Skill & A1 & A2 & A1 & A2 & $\Delta$ \\
\midrule
\multirow{2}{*}{ML}
 & Collaboration & 0.418 & 0.403 & 0.811 & 0.808 & $+0.399$ \\
 & Communication & 0.577 & 0.551 & 0.804 & 0.754 & $+0.215$ \\
\midrule
\multirow{2}{*}{SE}
 & Collaboration & 0.283 & 0.296 & 0.712 & 0.717 & $+0.425$ \\
 & Communication & 0.418 & 0.431 & 0.696 & 0.702 & $+0.275$ \\
\bottomrule
\end{tabular}
\end{table}

\textbf{Comparison with a fine-tuned BERT classifier.}
To position LLM-as-a-Judge against a contemporary supervised baseline, we reimplemented the BERT-based sequence classification architecture proposed by~\citet{Skondras_2023}, adding class-weighted cross-entropy to mitigate the strong class imbalance in our corpus. Both methods were evaluated on the subset of ML postings on which the two human annotators agreed (482 postings). BERT was trained and tested on a stratified 90/10 split, while LLM-as-a-Judge was evaluated on the same held-out test set without any fine-tuning. The full training configuration, per-class metrics, and confusion matrices are provided in Appendix~\ref{app:bert-detail}; Table~\ref{tab11} summarizes the headline results.

\begin{table}[t]
\centering
\caption{Performance of fine-tuned BERT and LLM-as-a-Judge on the human-agreed ground-truth test set. Per-class metrics and confusion matrices are reported in Appendix~\ref{app:bert-detail}.}
\label{tab11}
\small
\begin{tabular}{llccc}
\toprule
Method & Task & Accuracy & Macro-F1 & Weighted-F1 \\
\midrule
BERT & Collaboration & 0.92 & 0.64 & 0.91 \\
BERT & Communication & 0.84 & 0.77 & 0.83 \\
LLM-as-a-Judge & Collaboration & 0.96 & 0.86 & 0.96 \\
LLM-as-a-Judge & Communication & 0.88 & 0.86 & 0.88 \\
\bottomrule
\end{tabular}
\end{table}

LLM-as-a-Judge outperforms the supervised baseline on both tasks. For communication, accuracy improves from 0.84 to 0.88 and macro-F1 from 0.77 to 0.86; for collaboration, the gains are larger, with accuracy rising from 0.92 to 0.96 and macro-F1 from 0.64 to 0.86. The macro-F1 gap is the diagnostic result. BERT attains a high weighted-F1 on collaboration (0.91), almost entirely by predicting the majority class, correctly rejecting only one of the four negative postings in the test set, and the contrast between its high weighted-F1 and its low macro-F1 is the fingerprint of a classifier dominated by class imbalance. LLM-as-a-Judge, by contrast, keeps macro-F1 and weighted-F1 close to each other on both tasks, indicating that its predictions remain balanced across classes even when one class is heavily under-represented.

These results should be interpreted with appropriate care. The BERT configuration we evaluate is the architecture proposed in the original study; a larger labeled corpus, a different encoder, or a more aggressive sampling strategy could reasonably narrow the gap. The n-gram and BERT baselines together do not exhaust the space of existing approaches, but they bracket the two extremes most commonly used in practice. The broader takeaway is methodological. Supervised sequence classifiers remain attractive when large, balanced corpora are available, while prompt-based contextual evaluation becomes attractive when labels are scarce, imbalanced, or expensive to obtain, a pattern that is representative of much labor-market and SE research.

\subsection{Limitations of LLM-as-a-Judge}
\label{sec:llm-limitations}

The comparative study also surfaced limitations of LLM-as-a-Judge that practitioners should weigh against the gains documented above. The most consistent issue we observed is that the model identifies communication skills less reliably than collaboration skills. Even when the prompt was carefully refined, and the model avoided overt hallucination, i.e., it did not fabricate context that was absent from the job posting, it occasionally failed to apply fine-grained annotation rules consistently. As a concrete example, our prompt specified that a single mention of mentoring or guiding should be considered evidence of communication ability only when it was elaborated or accompanied by additional communication cues. The model applied this rule correctly in most cases, but in a non-trivial minority of job postings, it assigned a positive label to communication on the basis of an isolated, unelaborated mention of mentoring. This pattern indicates that current language models, even when prompted with detailed annotation guidelines, do not yet apply nuanced rule systems with the deterministic consistency that supervised annotation requires.

Two implications follow. First, LLM-as-a-Judge is best framed as an assistive annotation mechanism rather than a fully autonomous evaluator. The stochastic decoding process behind contemporary LLMs means that repeated runs on the same input can produce slightly different classifications, particularly when category boundaries are subtle. Any deployment that targets research-grade reliability should therefore include a human-in-the-loop validation layer, with humans focusing their limited attention on job postings flagged by the model as borderline or internally inconsistent. Second, the framework's reliability depends sensitively on the granularity of the prompt. Rules that humans can apply reflexively, such as the distinction between an isolated and an elaborated mention of mentoring, may need to be operationalized in the prompt with explicit examples and counterexamples before the model can follow them consistently.

In summary, LLM-as-a-Judge occupies a productive middle ground between lexical methods, which scale but lack contextual understanding, and supervised neural classifiers, which can model context but require labels that are costly to obtain. Used as a first-pass annotator whose output is reviewed or spot-checked by humans, it provides a favorable trade-off between annotation cost and reliability, and is particularly well suited to labor-market studies in which skill taxonomies evolve faster than supervised corpora can be re-labeled.

\subsection{Implications for Stakeholders}

Combining the empirical and methodological findings yields actionable implications for the four stakeholders.

\textbf{For educators}, the consistent demand for collaboration and communication across regions, industries, and seniority levels indicates that ML curricula should treat interpersonal development as a sustained thread of the degree rather than a single capstone elective. The U-shaped seniority profile is particularly relevant here. Graduates entering the labor market are expected to demonstrate interpersonal competence from day one, not to acquire it after acquiring technical proficiency. Curricular formats that support this include team-based capstone projects with non-technical stakeholders, structured presentation training, and explicit instruction in cross-functional collaboration. Our data also suggest that healthcare and industrial ML as growing application sectors place particularly strong demands on interpersonal capability, which may warrant targeted coursework for students intending to enter those sectors.

\textbf{For educators (facet-level).} The composition data give the curricular recommendation a sharper target. The competencies employers name most often are working across an expertise boundary, explaining analytical results to a named audience, and, from senior level onward, mentoring. Written communication is the notable asymmetry. It is comparatively rarely required in postings, but the sectors in which it is required, healthcare and industrial ML, are precisely the growth areas for applied ML, and documentation practice is independently necessary for reproducibility. We would therefore treat its low prevalence in advertisements as an argument for teaching it deliberately rather than as evidence that it can be omitted. Generic teamwork exercises address the least-demanded collaboration facet; interdisciplinary project work with a non-technical counterpart addresses the most-demanded one, and course designs of this kind have already been implemented and evaluated in practice~\citep{Azamnouri_2026_c}.

\textbf{For recruiters}, the high prevalence of implicit expression indicates that current job-advertisement practice often relies on indirect language to convey interpersonal requirements. This carries two costs. First, candidates who read postings literally may infer that interpersonal skills are not required, leading to skill-deficient applications and inflated screening costs downstream. Second, NLP-based applicant-tracking systems that rely on keyword matching will systematically fail to surface candidates who articulate interpersonal capability in non-standard terms. Both costs can be reduced by a more structured and explicit articulation of soft-skill requirements in job postings. Our taxonomy offers a starting point for this articulation.

\textbf{For recruiters (facet-level).} The specificity gap is directly actionable. A posting that names the audience and the channel communicates a requirement that candidates can prepare for and that screening can assess; a posting that asserts \enquote{excellent communication skills} does neither. Our facet inventory can serve as a checklist for this purpose: which counterparties does the role integrate with, to which audiences must results be explained, in which channel, and does the role carry mentoring or leadership responsibility. The role-level profiles also indicate that a single organizational template applied to all ML vacancies will misdescribe both ends of the distribution, overstating the interpersonal demands of infrastructure roles and understating those of client-facing architecture roles.

\textbf{For job seekers}, two seniority-specific signals stand out. First, the high entry-level demand means that early-career ML candidates should expect to demonstrate interpersonal capability immediately, not in some future career stage. Second, the renewed intensification at senior and management levels means that candidates planning advancement should treat interpersonal skill development as a precondition for promotion rather than a derivative of technical seniority. The mid-level dip in explicit communication requirements should not be read as a signal that communication matters less at that stage; the more plausible interpretation is that it is assumed.

\textbf{For job seekers (facet-level).} The role profiles are more informative for career planning than the aggregate figures. Candidates moving from an infrastructure-oriented role towards architecture, consulting, or management are moving into the part of the ML labor market where stakeholder coordination and explanation are demanded several times as often (stakeholder coordination in 9.1\% of MLOps postings against 40.7\% of management and 67.9\% of solutions-architect postings), and the transition from mid to senior level is characterized specifically by the appearance of mentoring and leadership requirements rather than by a general rise in interpersonal expectations.

\textbf{For researchers}, the methodological framework offers a transferable template. The combination of stratified multi-country sampling, BERTScore-routed translation, and hybrid human–LLM annotation can be applied beyond ML and SE to any technical labor market for which multilingual postings are accessible. The framework also supports longitudinal extensions. Re-running the pipeline annually would enable tracking of how soft-skill demand evolves as the ML profession matures and as the broader labor market responds to automation.

\textbf{For researchers (facet-level).} The facet pass is a lightweight, reusable second stage for any pipeline that produces binary skill labels. It requires no additional annotation, it is deterministic and auditable, and it converts a prevalence study into a compositional one. Our results suggest this is where the analytical value lies. The aggregate prevalences in our corpus replicate what prior work leads one to expect, whereas the compositional differences across roles, seniority tiers, and domains were not predictable from that literature and in one case (industry) are invisible without it.

\section{Threats to Validity}
\label{sec:threats-to-validity}

We classify the threats to validity affecting our study along the four standard dimensions (internal, external, construct, and conclusion validity). For each threat, we identify the source of risk, describe the mitigations applied during the study design, and acknowledge the residual limitations that should be considered when interpreting the results.

\subsection{Internal Validity}

The study relies on open-weight LLMs whose performance constrains the upper bound of the annotation pipeline. More capable proprietary models, such as the latest commercial frontier systems, may yield different or improved agreement with human annotators. We mitigated this threat by evaluating four open-weight candidates against a high-agreement human ground truth and selecting the best-performing model for each target skill. The residual risk is that an alternative configuration based on closed-source models might shift the LLM–human Cohen's $\kappa$ values, although the human-centered majority-vote design limits the downstream impact of such shifts on the final labels.

The translation pipeline was validated automatically through BERTScore and manually for the Vietnamese and German subsets. No professional translators or native speakers were available to verify the Japanese, French, Portuguese, or Spanish translations, leaving residual uncertainty about translation fidelity for these languages. The BERTScore-based routing logic provides a per-instance semantic-similarity guarantee against a 0.85 threshold, which partially compensates for the absence of human review, but it cannot fully substitute for native-speaker verification on domain-specific terminology.

Using SerpApi as a regulated aggregator for Google Jobs provided broad and compliant coverage of multiple recruitment platforms but introduced several data-quality risks. First, duplication arises when employers post identical vacancies across multiple boards; we mitigated this through a three-tier detection procedure combining exact-match removal, fuzzy matching, and semantic similarity using BERT embeddings, with a 90\% threshold and manual review of flagged pairs. Second, truncation occurs when third-party agencies surface only partial descriptions in Google Jobs, which can suppress the textual evidence required for accurate skill classification; such cases were identified during the human annotation pass and replaced through targeted manual scraping where feasible. Third, abbreviation ambiguity in queries, most notably the term \enquote{ML}, can return false positives unrelated to machine learning; manual filtering during cleaning addressed these cases. Although the preprocessing pipeline substantially reduces these risks, the residual concern is that automated job-market data cannot be treated as reliable by default. Our results should be interpreted with this caveat in mind, and the broader methodological lesson is that API-based collection requires careful preprocessing rather than direct downstream use.

The role dimension analyzed in Section~\ref{sec:rq2-role} rests on a rule-based classification of job titles, manually verified by the annotators. Job titles are an imperfect proxy for the work a role actually involves; titles are not standardized across organizations or regions, and the categories are unevenly populated, with the two smallest groups comprising 25 and 28 postings. Role is also correlated with seniority by construction, so the role and seniority associations we report are marginal rather than mutually adjusted; a multivariate model would be required to separate them, which the subgroup sizes in our corpus do not support.

\subsection{External Validity}

The corpus constitutes a cross-sectional snapshot. All postings were collected through SerpApi within a single time window, so the longitudinal stability of the reported prevalences is not directly tested and should not be assumed. Moreover, the stratified sample spans five continents but does not include Oceania. The absence of data from this region restricts the claim of global representativeness and may introduce regional bias. Within the covered continents, country-level coverage is also uneven, reflecting differences in the volume of postings available through SerpApi for each language–region combination.

While the overall corpus of 1,000 job postings supports robust aggregate analysis, the subgroup sample sizes used for RQ2 vary substantially. The management seniority subgroup (n=30), the industrial subgroup (n=33), and the healthcare subgroup (n=37) are all small relative to the senior (n=321) and technology (n=383) subgroups. The corresponding 95\% confidence intervals are therefore wide, and subgroup-level comparisons should be treated as descriptive rather than as the basis for strong inferential claims. We pooled the two smallest industry labels, manufacturing and energy, into a single Industrial category to avoid spurious estimates, but the underlying imbalance remains.

Translation introduces variability in how interpersonal competencies are expressed across linguistic and cultural contexts. Idiomatic phrases that signal collaboration or communication in one language may translate to English formulations that the LLM judges differently, and recruitment conventions themselves differ across regions. The findings should therefore be interpreted as describing the English-translated representations of the original postings rather than the original-language constructs directly. Transferability to postings written in languages outside our coverage, or to industries and regions underrepresented in our sample, cannot be assumed.

\subsection{Construct Validity}

The collaboration and communication taxonomies were derived from prior work~\citep{Khaouja_2019, Pol_kov_2023} and refined by inter-annotator agreement during the labeling phase. Despite this grounding, the taxonomy may not exhaust the full range of expressions associated with the two constructs. Idiosyncratic phrasings, region-specific idioms, or emerging vocabulary may fall outside the taxonomy and remain unlabeled. The hybrid LLM-as-a-Judge architecture partially mitigates this risk because the LLM can reason over context rather than match fixed vocabulary, but the human annotators' reference taxonomy still bounds what the pipeline is trained to detect. A more expansive taxonomy incorporating additional synonyms and indirect expressions could yield somewhat higher recall.

The distinction between explicit and implicit collaboration and communication skills is operationalized by comparing the n-gram baseline to the hybrid human–LLM ground truth. This operationalization defines explicit as a match to the predefined regular expression patterns, and implicit as any other positive label. A different choice of explicit-mention patterns would shift the boundary between the two categories; the implicit-share estimates should therefore be read as bounded by the chosen pattern set rather than as absolute measurements.

The facet coding introduces a further construct validity consideration. Unlike the binary skill labels, the facet codes have no independently annotated ground truth; they are a deterministic re-reading of postings whose labels were established by the hybrid pipeline. We tuned the rule set for precision and audited it on a random sample of evidence clauses, obtaining approximately 94\% precision, but recall is deliberately not maximized. Between 6.6\% and 10.2\% of positively labeled postings, depending on skill and domain, could not be attributed to any facet, and the reported facet shares are therefore lower bounds. A more permissive pattern set would raise all shares somewhat; we would not expect it to reverse the orderings, since the contrasts we interpret (cross-functional against teamwork, ML against SE, role against role) rest on differences well outside the plausible range of the residual. A second consideration concerns the boundary between stakeholder coordination and audience-facing interaction, which our rules draw by the governing verb. Alternative conventions would redistribute weight between these two facets, though not between collaboration and communication as constructs, since the same clause supports one of them either way.

\subsection{Conclusion Validity}

The classification metrics used to evaluate the n-gram baseline (precision, recall, F1, Cohen's $\kappa$) can be sensitive to class imbalance, which is present in our corpus. 88.6\% of ML job postings and 91.4\% of SE job postings are positive for collaboration. Under such an imbalance, small annotation perturbations can produce non-trivial shifts in $\kappa$ in particular. We mitigated this risk through the high inter-human agreement achieved during annotation ($\kappa \geq 0.919$ on the ML subset and $\kappa \geq 0.972$ on the SE subset), which limits the magnitude of plausible perturbations, but the residual sensitivity should be borne in mind when comparing performance across subsets.

The hybrid human–LLM ground truth depends on the consistent application of the annotation criteria across both human annotators. Although Cohen's $\kappa$ values place inter-human agreement in the almost-perfect range, residual subjectivity in borderline cases, particularly those involving implicit expressions, cannot be eliminated. The LLM review step in our pipeline addressed cases where the LLM flagged evidence overlooked by the human annotators, but neither human nor LLM judgment is infallible on contextually ambiguous job postings.

Our empirical comparison covers an n-gram baseline and a fine-tuned BERT classifier as supervised reference points. We did not run head-to-head comparisons on the same corpus against embedding-based pipelines, e.g., Word\-2Vec or DBpedia-augmented extractors~\citep{Khaouja_2019, Calanca_2019}, or against SkillSpan-style neural sequence labeling~\citep{Zhang_2022}. Consequently, our results do not support categorical claims that the proposed pipeline outperforms the full space of available alternatives, only that it substantially improves on the keyword baseline and the standard supervised BERT baseline most commonly cited in prior work. A direct benchmark against neural span-extraction methods on this corpus is a natural next step and would situate the hybrid pipeline more precisely within the broader skill-extraction landscape.

The facet analysis multiplies the number of hypothesis tests substantially, with twelve facet tests per contextual dimension and twelve for the domain contrast. We control the family-wise error rate with Holm--Bonferroni adjustment within each family and report Cram\'er's $V$ throughout, and we confine our substantive interpretation to associations that survive adjustment. Facet analyses are additionally conditioned on a positive skill label, which is itself an outcome of the annotation pipeline; the compositional shares should accordingly be read as descriptions of how requirements are expressed given that they are expressed, not as unconditional population quantities.

Taken together, these threats bound rather than undermined the contribution. The empirical findings are best read as a robust characterization of the labor-market segment our sample captures, at the point in time at which it was collected, and as a foundation for replication in adjacent domains, regions, and periods; we would not claim that the precise prevalence figures constitute the final word on collaboration and communication demand in ML hiring. The methodological framework, by contrast, transfers directly. The sampling, translation-routing, and hybrid human--LLM annotation pipeline can be reapplied wherever multilingual posting data are accessible.

\section{Conclusion}
\label{sec:conclusion}

This study examined the demand for collaboration and communication skills in ML job postings, comparing variation across regions, industries, and seniority levels and benchmarking ML against SE. To support the analysis, we developed an end-to-end pipeline combining semi-automated data collection via SerpApi, multilingual translation with BERTScore-based quality estimation and routing, and a hybrid annotation architecture in which an LLM-as-a-Judge serves as a third annotator alongside two human annotators.

The empirical findings establish that collaboration and communication are core requirements in ML hiring rather than peripheral attributes. At least one of the two skills was demanded in 92.0\% of ML job postings, with collaboration (88.6\%) more prevalent than communication (72.2\%); both proportions significantly exceed a 50\% benchmark. The baseline demand is high in every subgroup examined, while the intensity varies, with regional differences in communication the only variation that remains significant after correction for multiple testing. North American labor markets, healthcare, and industrial sectors, and senior and management positions display an explicit emphasis. The cross-domain comparison reveals that SE job postings show only marginally higher prevalence than ML job postings, with differences falling within the overlap of the corresponding confidence intervals. This near parity suggests that the slightly greater visibility of collaboration and communication skills requirements in SE reflects more standardized job-description conventions rather than a genuinely higher workplace demand. A further finding is that roughly one quarter of communication requirements and one fifth of collaboration requirements are expressed implicitly and would be missed by traditional keyword-based pipelines, underscoring the value of contextual annotation methods.

Decomposing these requirements into the facets of our taxonomy adds a compositional layer to this picture. Collaboration is demanded chiefly as integration across role boundaries rather than as team membership; communication is dominated by generic assertions, with fewer than half of the relevant postings naming an audience and only 18.6\% requiring written communication; the facets that intensify with seniority are specifically those presupposing responsibility for others; and the near parity between ML and SE holds for the level of demand but not for its orientation, which is outward-facing in ML and inward-facing in SE. Within ML, the advertised role differentiates interpersonal profiles more strongly than region, industry, seniority, or the comparison with SE.

These findings carry direct implications for stakeholders. Recruiters could benefit from more structured and explicit articulation of interpersonal requirements in ML job advertisements. Educators should integrate sustained opportunities for students to develop collaboration, presentation, and interdisciplinary communication skills alongside technical training. Job seekers, particularly those targeting senior or managerial positions, should recognize that career advancement in technical domains depends not only on technical proficiency but also on demonstrable interpersonal capability.

Overall, the results establish that collaboration and communication skills are fundamental employability requirements in ML job positions, with importance that persists across geographic, industrial, and organizational contexts and approaches the levels observed in the more mature SE profession. By combining substantive evidence on the human dimensions of ML work with a reproducible methodological framework, this paper offers practical guidance to educators, recruiters, and job seekers and lays a foundation for future research into the evolving role of collaboration and communication skills in technical employment markets.

Several directions remain for future work. The analysis could be extended to a broader set of soft skills, including adaptability, critical thinking, and leadership, and to additional regions such as Oceania and adjacent technical occupations such as MLOps and AI research engineering. Methodologically, domain-specific fine-tuning, retrieval-augmented prompting, and head-to-head comparison against neural span-extraction methods would refine and contextualize the present contribution. Finally, linking soft-skill demand to downstream labor-market outcomes such as salary, time-to-hire, and career progression would move the analysis from description toward causal and predictive understanding.

\appendix

\section{Facet Levels}
\begin{figure}
\includegraphics[width=1\textwidth]{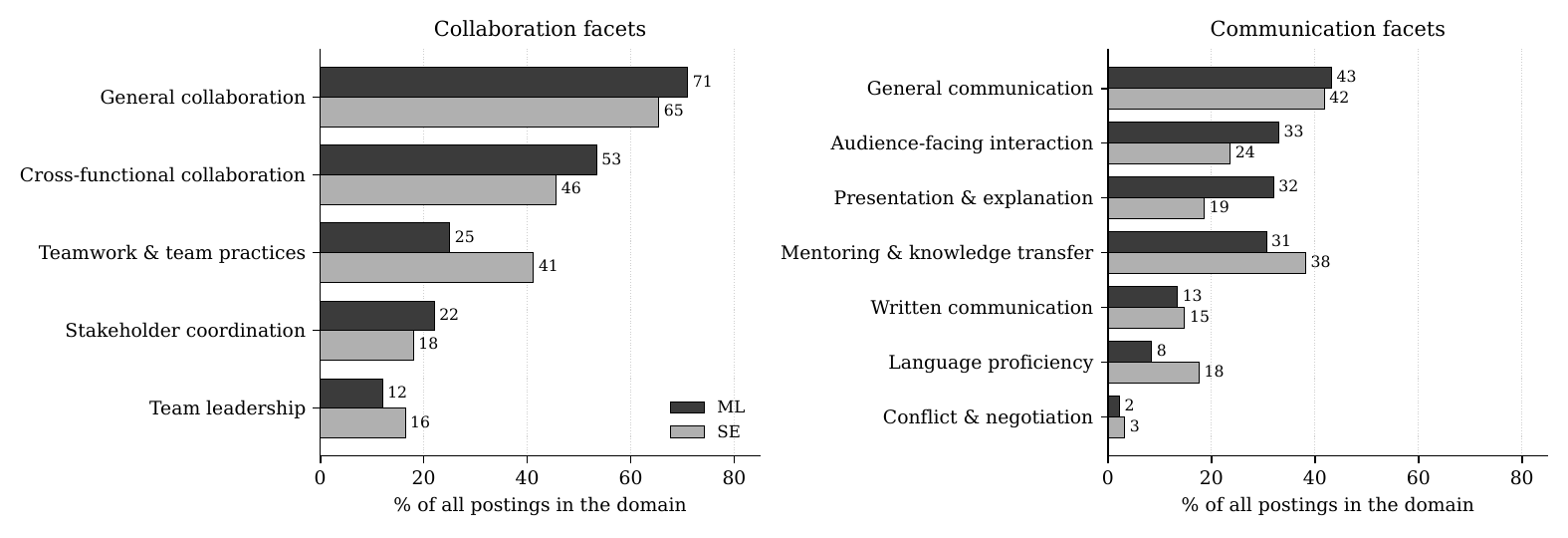}
\caption{Facet composition of collaboration and communication requirements in ML and SE job postings, as a percentage of all 500 postings per domain}
\label{facet-composition}
\end{figure}

\begin{figure}
\includegraphics[width=1\textwidth]{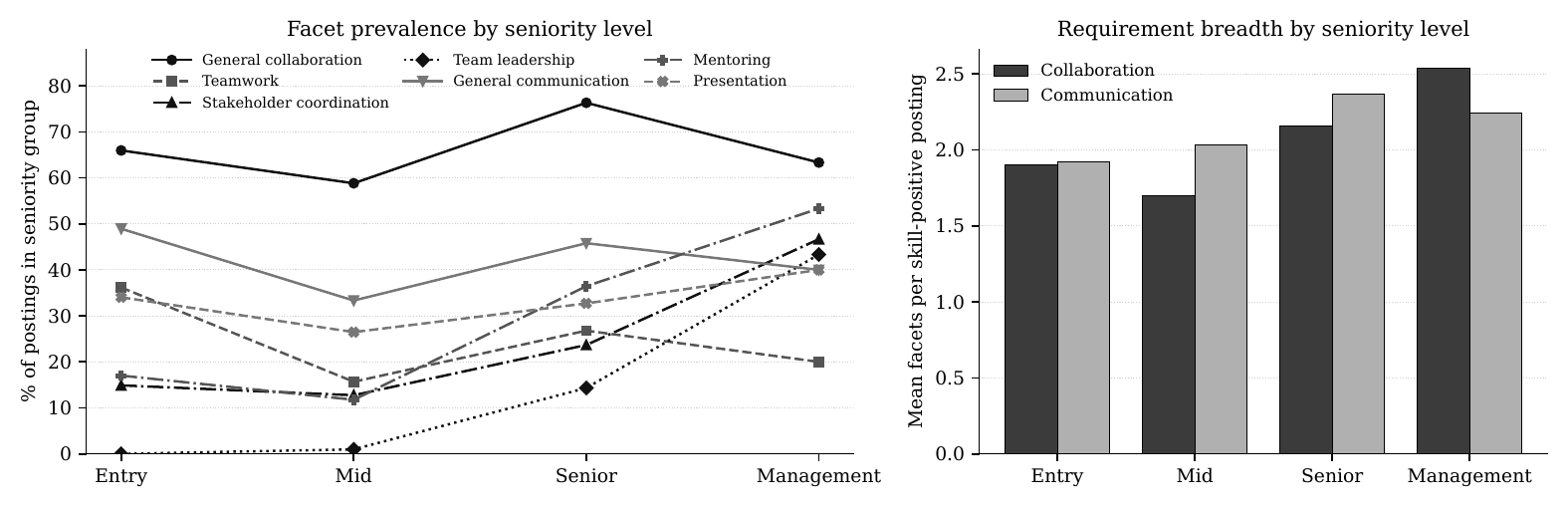}
\caption{Facet prevalence and requirement breadth by seniority level in the ML corpus. Left: percentage of postings in each seniority group expressing the facet. Right: mean number of distinct facets expressed per posting that requires the corresponding skill}
\label{facet-by-seniority}
\end{figure}

\begin{figure}
\centering
\includegraphics[width=0.95\textwidth]{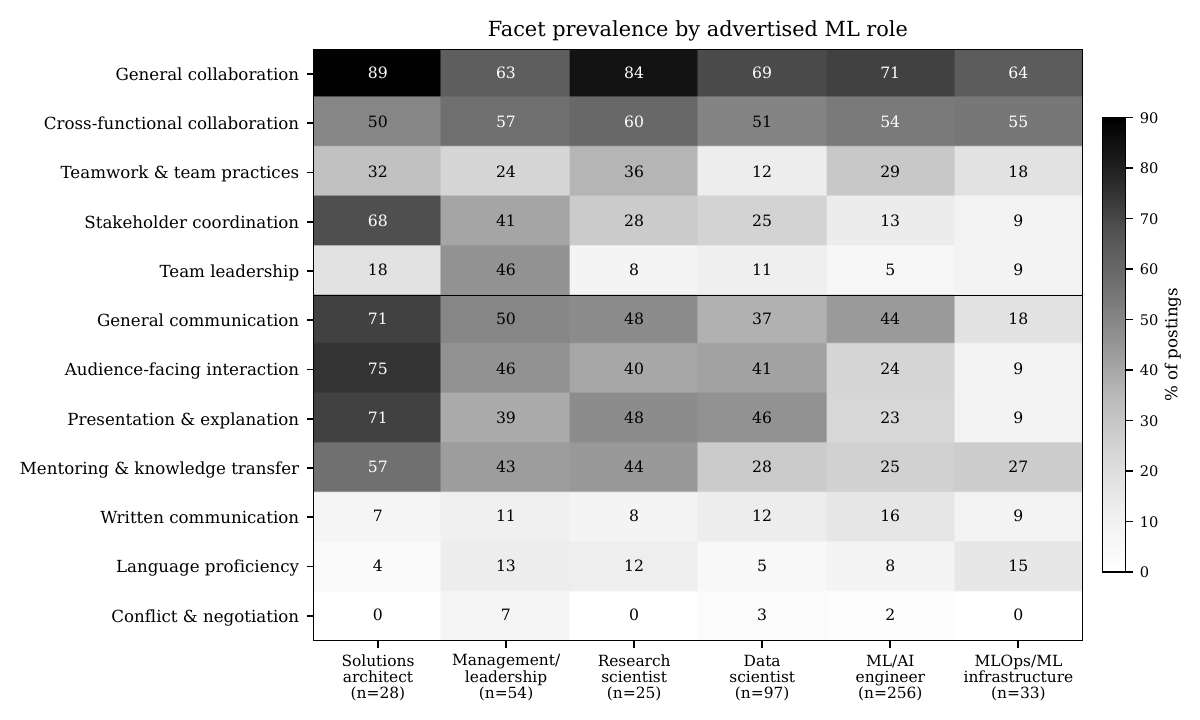}
\caption{Facet prevalence by advertised ML role, as a percentage of the postings in each role group. Collaboration facets above the rule, communication facets below}
\label{facet-by-role}
\end{figure}

This appendix collects the facet-level tables and figures referenced throughout Section~\ref{sec:results}. Table~\ref{tab:facet-composition} reports the facet composition of both constructs in the ML and SE corpora, and Fig.~\ref{facet-composition} visualizes the same contrast; Tables~\ref{tab:facet-seniority} and~\ref{tab:facet-region} disaggregate the ML corpus by seniority level and geographic region, with the seniority pattern also shown in Fig.~\ref{facet-by-seniority} and the role-level profiles in Fig.~\ref{facet-by-role}; Table~\ref{tab:facet-implicit} splits the facet shares by whether the underlying requirement is detected by the n-gram baseline; and Table~\ref{tab:app-facet-tests} consolidates the complete set of facet-level subgroup tests across all four contextual dimensions. Unless stated otherwise in the individual captions, percentages are shares of all postings in the respective group rather than of the skill-positive subset, $p$-values are Holm--Bonferroni-adjusted within each family of twelve facet tests, and the reported shares should be read as lower bounds, since the facet coder of Section~\ref{sec:facet-coding} is tuned for precision.

\begin{table}[t]
\centering
\caption{Facet composition of collaboration and communication requirements in ML and SE postings. The ML and SE columns give the share (\%) of all 500 postings in the domain that express the facet, and $\Delta$ their difference in percentage points. Adjusted $p$-values are from $\chi^2$ tests of independence with Holm--Bonferroni correction across the twelve facets; $V$ is Cram\'er's $V$.}
\label{tab:facet-composition}
\small
\begin{tabular}{@{}lrrrrr@{}}
\toprule
Facet & ML & SE & $\Delta$ & adj. $p$ & $V$ \\
\midrule
\multicolumn{6}{@{}l}{\emph{Collaboration}}\\
\quad General collaboration & 71.0 & 65.4 & $+5.6$ & 0.286 & 0.06 \\
\quad Cross-functional collaboration & 53.4 & 45.6 & $+7.8$ & 0.095 & 0.08 \\
\quad Teamwork \& team practices & 25.0 & 41.2 & $-16.2$ & $<$0.0001 & 0.17 \\
\quad Stakeholder coordination & 22.0 & 18.0 & $+4.0$ & 0.455 & 0.05 \\
\quad Team leadership & 12.0 & 16.4 & $-4.4$ & 0.277 & 0.06 \\
\addlinespace
\multicolumn{6}{@{}l}{\emph{Communication}}\\
\quad General communication & 43.2 & 41.8 & $+1.4$ & 1.000 & 0.01 \\
\quad Audience-facing interaction & 33.0 & 23.6 & $+9.4$ & 0.009 & 0.10 \\
\quad Presentation \& explanation & 32.0 & 18.6 & $+13.4$ & $<$0.0001 & 0.15 \\
\quad Mentoring \& knowledge transfer & 30.6 & 38.2 & $-7.6$ & 0.091 & 0.08 \\
\quad Written communication & 13.4 & 14.8 & $-1.4$ & 1.000 & 0.02 \\
\quad Language proficiency & 8.4 & 17.6 & $-9.2$ & 0.0002 & 0.14 \\
\quad Conflict \& negotiation & 2.2 & 3.2 & $-1.0$ & 0.988 & 0.03 \\
\midrule
Unattributed (collaboration) & 7.2 & 6.4 & $+0.8$ & --- & --- \\
Unattributed (communication) & 4.8 & 7.8 & $-3.0$ & --- & --- \\
\bottomrule
\end{tabular}
\end{table}
 
\begin{table}[t]
\centering
\caption{Facet prevalence by seniority level in the ML corpus (\% of postings in the seniority group). $p$-values from $\chi^2$ tests of independence, verified by Monte-Carlo resampling with exact margins ($2\times10^{5}$ resamples) where an expected cell count falls below five, and adjusted with Holm--Bonferroni across the twelve facets.}
\label{tab:facet-seniority}
\small
\footnotesize
\setlength{\tabcolsep}{3pt}
\begin{tabular}{@{}lrrrrrr@{}}
\toprule
 & \multicolumn{4}{c}{\% of postings} & & \\
\cmidrule(lr){2-5}
Facet & Entry & Mid & Senior & Mgmt & adj. $p$ & $V$ \\
 & ($n$=47) & ($n$=102) & ($n$=321) & ($n$=30) & & \\
\midrule
\multicolumn{7}{@{}l}{\emph{Collaboration}}\\
\quad General collaboration & 66.0 & 58.8 & 76.3 & 63.3 & 0.038 & 0.16 \\
\quad Cross-functional collaboration & 48.9 & 48.0 & 54.8 & 63.3 & 1.000 & 0.08 \\
\quad Teamwork \& team practices & 36.2 & 15.7 & 26.8 & 20.0 & 0.257 & 0.13 \\
\quad Stakeholder coordination & 14.9 & 12.7 & 23.7 & 46.7 & 0.005 & 0.19 \\
\quad Team leadership & 0.0 & 1.0 & 14.3 & 43.3 & $<$0.0001 & 0.31 \\
\addlinespace
\multicolumn{7}{@{}l}{\emph{Communication}}\\
\quad General communication & 48.9 & 33.3 & 45.8 & 40.0 & 0.897 & 0.11 \\
\quad Audience-facing interaction & 23.4 & 28.4 & 35.5 & 36.7 & 1.000 & 0.09 \\
\quad Presentation \& explanation & 34.0 & 26.5 & 32.7 & 40.0 & 1.000 & 0.07 \\
\quad Mentoring \& knowledge transfer & 17.0 & 11.8 & 36.4 & 53.3 & $<$0.0001 & 0.26 \\
\quad Written communication & 19.1 & 11.8 & 14.0 & 3.3 & 1.000 & 0.09 \\
\quad Language proficiency & 8.5 & 10.8 & 7.5 & 10.0 & 1.000 & 0.05 \\
\quad Conflict \& negotiation & 0.0 & 2.9 & 2.2 & 3.3 & 1.000 & 0.05 \\
\bottomrule
\end{tabular}
\end{table}
 
\begin{table}[t]
\centering
\caption{Facet prevalence by geographic region in the ML corpus (\% of the 100 postings per region), with Holm--Bonferroni-adjusted $\chi^2$ tests across the twelve facets.}
\label{tab:facet-region}
\small
\footnotesize
\setlength{\tabcolsep}{3pt}
\begin{tabular}{@{}lrrrrrrr@{}}
\toprule
 & \multicolumn{5}{c}{\% of postings} & & \\
\cmidrule(lr){2-6}
Facet & Afr. & Asia & Eur. & N.Am. & S.Am. & adj. $p$ & $V$ \\
\midrule
\multicolumn{8}{@{}l}{\emph{Collaboration}}\\
\quad General collaboration & 80 & 68 & 63 & 79 & 65 & 0.060 & 0.16 \\
\quad Cross-functional collaboration & 66 & 59 & 41 & 62 & 39 & 0.0004 & 0.22 \\
\quad Teamwork \& team practices & 32 & 24 & 27 & 30 & 12 & 0.051 & 0.16 \\
\quad Stakeholder coordination & 23 & 17 & 15 & 35 & 20 & 0.036 & 0.17 \\
\quad Team leadership & 11 & 15 & 8 & 15 & 11 & 0.689 & 0.08 \\
\addlinespace
\multicolumn{8}{@{}l}{\emph{Communication}}\\
\quad General communication & 60 & 33 & 35 & 55 & 33 & $<$0.0001 & 0.24 \\
\quad Audience-facing interaction & 57 & 21 & 22 & 42 & 23 & $<$0.0001 & 0.30 \\
\quad Presentation \& explanation & 57 & 23 & 19 & 40 & 21 & $<$0.0001 & 0.31 \\
\quad Mentoring \& knowledge transfer & 46 & 18 & 26 & 37 & 26 & 0.001 & 0.21 \\
\quad Written communication & 13 & 8 & 15 & 19 & 12 & 0.689 & 0.11 \\
\quad Language proficiency & 0 & 9 & 20 & 4 & 9 & $<$0.0001 & 0.24 \\
\quad Conflict \& negotiation & 2 & 1 & 1 & 5 & 2 & 0.689 & 0.10 \\
\bottomrule
\end{tabular}
\end{table}
 
\begin{table}[t]
\centering
\caption{Facet composition of implicitly and explicitly expressed requirements. Columns give the percentage of skill-positive postings carrying each facet, split by whether the posting is detected by the n-gram baseline (explicit) or not (implicit-only).}
\label{tab:facet-implicit}
\small
\setlength{\tabcolsep}{5pt}
\begin{tabular}{@{}lrrrr@{}}
\toprule
 & \multicolumn{2}{c}{ML} & \multicolumn{2}{c}{SE} \\
\cmidrule(lr){2-3}\cmidrule(lr){4-5}
Facet & Implicit & Explicit & Implicit & Explicit \\
\midrule
\multicolumn{5}{@{}l}{\emph{Collaboration}}\\
\quad General collaboration & 22.9 & 96.0 & 16.7 & 93.8 \\
\quad Cross-functional collaboration & 29.2 & 68.9 & 31.8 & 57.2 \\
\quad Teamwork \& team practices & 20.8 & 30.3 & 35.6 & 48.9 \\
\quad Stakeholder coordination & 19.8 & 26.2 & 15.9 & 21.2 \\
\quad Team leadership & 11.5 & 14.1 & 19.7 & 17.2 \\
\addlinespace
\multicolumn{5}{@{}l}{\emph{Communication}}\\
\quad General communication & 4.0 & 80.9 & 6.5 & 77.9 \\
\quad Audience-facing interaction & 24.2 & 53.8 & 17.9 & 37.2 \\
\quad Presentation \& explanation & 29.3 & 50.0 & 14.6 & 29.1 \\
\quad Mentoring \& knowledge transfer & 51.5 & 38.9 & 46.3 & 51.9 \\
\quad Written communication & 11.1 & 21.4 & 15.4 & 21.3 \\
\quad Language proficiency & 19.2 & 8.8 & 24.4 & 22.5 \\
\quad Conflict \& negotiation & 0.0 & 4.2 & 4.1 & 4.3 \\
\bottomrule
\end{tabular}
\end{table}
 
\begin{table}[t]
\centering
\caption{Complete set of facet-level subgroup tests on the ML corpus. Each cell reports the Holm--Bonferroni-adjusted $p$-value and Cram\'er's $V$ of a $\chi^2$ test of independence between the contextual dimension and the facet indicator; adjustment is applied within each dimension (twelve tests per family).}
\label{tab:app-facet-tests}
\small
\footnotesize
\setlength{\tabcolsep}{2.5pt}
\begin{tabular}{@{}lrrrrrrrr@{}}
\toprule
 & \multicolumn{2}{c}{Seniority} & \multicolumn{2}{c}{Region} & \multicolumn{2}{c}{Industry} & \multicolumn{2}{c}{Role} \\
\cmidrule(lr){2-3}\cmidrule(lr){4-5}\cmidrule(lr){6-7}\cmidrule(lr){8-9}
Facet & $p$ & $V$ & $p$ & $V$ & $p$ & $V$ & $p$ & $V$ \\
\midrule
\multicolumn{9}{@{}l}{\emph{Collaboration}}\\
\quad General collaboration & 0.038 & 0.16 & 0.060 & 0.16 & 1.000 & 0.08 & 0.462 & 0.14 \\
\quad Cross-functional collaboration & 1.000 & 0.08 & 0.0004 & 0.22 & 1.000 & 0.08 & 1.000 & 0.05 \\
\quad Teamwork \& team practices & 0.257 & 0.13 & 0.051 & 0.16 & 0.194 & 0.14 & 0.112 & 0.17 \\
\quad Stakeholder coordination & 0.005 & 0.19 & 0.036 & 0.17 & 1.000 & 0.03 & $<$0.0001 & 0.35 \\
\quad Team leadership & $<$0.0001 & 0.31 & 0.689 & 0.08 & 0.533 & 0.12 & $<$0.0001 & 0.38 \\
\addlinespace
\multicolumn{9}{@{}l}{\emph{Communication}}\\
\quad General communication & 0.897 & 0.11 & $<$0.0001 & 0.24 & 1.000 & 0.10 & 0.009 & 0.20 \\
\quad Audience-facing interaction & 1.000 & 0.09 & $<$0.0001 & 0.30 & 1.000 & 0.03 & $<$0.0001 & 0.31 \\
\quad Presentation \& explanation & 1.000 & 0.07 & $<$0.0001 & 0.31 & 1.000 & 0.06 & $<$0.0001 & 0.32 \\
\quad Mentoring \& knowledge transfer & $<$0.0001 & 0.26 & 0.001 & 0.21 & 1.000 & 0.07 & 0.014 & 0.20 \\
\quad Written communication & 1.000 & 0.09 & 0.689 & 0.11 & 0.002 & 0.20 & 1.000 & 0.09 \\
\quad Language proficiency & 1.000 & 0.05 & $<$0.0001 & 0.24 & 0.920 & 0.11 & 0.936 & 0.11 \\
\quad Conflict \& negotiation & 1.000 & 0.05 & 0.689 & 0.10 & 1.000 & 0.07 & 0.462 & 0.14 \\
\bottomrule
\end{tabular}
\end{table}

\section{Translation Routing Outcomes}
\label{app:translation}

Table~\ref{tab:translation-routing} summarizes translation quality and routing outcomes by language group and field type for the pipeline described in Section~\ref{sec:methodology}. Three regimes emerge. For typologically distant languages (Vietnamese, Japanese), noisy fields such as location frequently fail the fidelity threshold for both primary models, so the fallback handles the majority of these cases (53.8\%). For Portuguese/Spanish and German/French, routing is efficiency-driven. The 8B model suffices for short, formulaic fields. Sauerkraut is preferred for longer descriptions, where it offers a higher mean score and lower variance, and the fallback is rarely required (5.1\% and 8.0\% of location instances, respectively). Across all groups, a consistent text-type hierarchy holds. Titles are handled well by small models, locations expose threshold dynamics and data noise, and descriptions benefit from larger capacity. Because selection is governed by per-instance sufficiency rather than average performance, a model with a lower mean BERTScore can nevertheless be selected most frequently; for Portuguese/Spanish location strings, for example, Sauerkraut won 47 of 50 head-to-head comparisons, yet the 8B model was routed in 48.0\% of cases because its translations already cleared the fidelity floor.

\begin{table}[t]
\centering
\caption{Translation quality and routing outcomes by language group and field: mean BERTScore F1 of the two primary models and the model selected most frequently by the router. The fallback (Llama-3.1-70B) is invoked only when both primary candidates score below the 0.85 threshold.}
\label{tab:translation-routing}
\small
\setlength{\tabcolsep}{4pt}
\begin{tabular}{@{}llccl@{}}
\toprule
 & & \multicolumn{2}{c}{Mean BERTScore F1} & \\
\cmidrule(lr){3-4}
Language group & Field & 8B & Sauerkr. & Most selected (share) \\
\midrule
\multirow{3}{*}{Vietnamese/Japanese}
 & Title       & 0.937 & 0.937 & 8B (100.0\%) \\
 & Location    & 0.832 & 0.849 & Fallback (53.8\%) \\
 & Description & 0.853 & 0.878 & Sauerkraut (75.0\%) \\
\midrule
\multirow{3}{*}{Portuguese/Spanish}
 & Title       & 0.977 & 0.980 & 8B (72.7\%) \\
 & Location    & 0.938 & 0.961 & 8B (48.0\%) \\
 & Description & 0.918 & 0.919 & Sauerkraut (53.3\%) \\
\midrule
\multirow{3}{*}{German/French}
 & Title       & 0.941 & 0.946 & 8B (58.8\%) \\
 & Location    & 0.952 & 0.960 & 8B (82.7\%) \\
 & Description & 0.863 & 0.889 & Sauerkraut (52.8\%) \\
\bottomrule
\end{tabular}
\end{table}

\section{Detailed Comparison of Fine-tuned BERT and LLM-as-a-Judge}
\label{app:bert-detail}

This appendix reports the training configuration, per-class metrics, and confusion matrices for the comparison summarized in Section~\ref{sec:method-comparison}.

\subsection{Training Configuration}
\label{app:bert-config}

We reimplemented the BERT-based sequence classification architecture of~\citet{Skondras_2023}, introducing a single modification to mitigate the strong class imbalance in our corpus. The evaluation set was constructed from the subset of ML postings on which both human annotators agreed, yielding 482 postings with high-quality supervision. We adopted the original training configuration: the bert-base-uncased encoder, a maximum input length of 256 tokens, a learning rate of $2 \times 10^{-5}$, a batch size of 8, three training epochs, and a stratified 90/10 train-test split, and added class-weighted cross-entropy to penalize errors on the underrepresented class. The LLM-as-a-Judge framework was evaluated on the same held-out test set without any additional fine-tuning.

\subsection{Per-Class Results for LLM-as-a-Judge}
\label{app:llm-per-class}

\begin{figure}[t]
\centering
\includegraphics[width=0.5\textwidth]{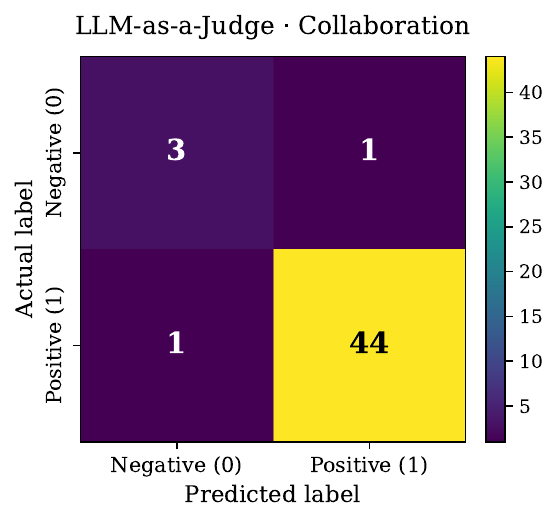}
\caption{Confusion matrices for LLM-as-a-Judge on the human-agreed test set -- collaboration.}
\label{confusion-matrices-llm-collaboration}
\end{figure}

\begin{figure}[t]
\centering
\includegraphics[width=0.5\textwidth]{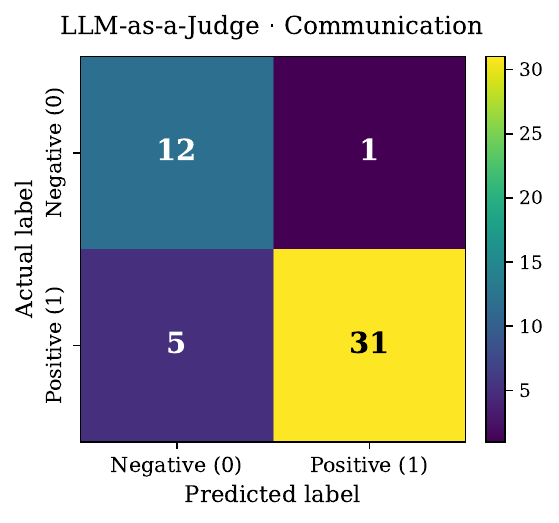}
\caption{Confusion matrices for LLM-as-a-Judge on the human-agreed test set -- communication.}
\label{confusion-matrices-llm-communication}
\end{figure}

Table~\ref{tab:app-per-class} reports the per-class precision, recall, and F1 scores for both methods on both tasks; Figs.~\ref{confusion-matrices-llm-collaboration} and~\ref{confusion-matrices-llm-communication} show the corresponding confusion matrices for LLM-as-a-Judge.

On collaboration, performance is both strong and symmetric. The positive class is identified with near-perfect scores (precision, recall, and F1 all equal to 0.98), while precision and recall on the negative class are both 0.75, computed over only four supporting instances. The confusion matrix in Fig.~\ref{confusion-matrices-llm-collaboration} shows that misclassifications are rare and that the predictions remain well calibrated despite the small size of the negative class.

On communication, the class-wise breakdown shows that the model is highly precise on the positive class (precision of 0.97, recall of 0.86, F1 of 0.91). When it asserts that a job posting requires communication ability, it is almost always right, but it misses five positive instances. The negative class is identified with a precision of 0.71 and a recall of 0.92, indicating that the framework adopts a conservative stance toward asserting the presence of communication skills. The confusion matrix in Fig.~\ref{confusion-matrices-llm-communication} corroborates this reading.

\begin{table}[t]
\centering
\caption{Per-class precision (P), recall (R), and F1 on the human-agreed test set. Support denotes the number of test instances per class.}
\label{tab:app-per-class}
\small
\begin{tabular}{lllcccc}
\toprule
Method & Task & Class & P & R & F1 & Support \\
\midrule
\multirow{4}{*}{LLM-as-a-Judge}
 & \multirow{2}{*}{Collaboration} & Positive & 0.98 & 0.98 & 0.98 & 45 \\
 &                                & Negative & 0.75 & 0.75 & 0.75 & 4  \\
 & \multirow{2}{*}{Communication} & Positive & 0.97 & 0.86 & 0.91 & 36 \\
 &                                & Negative & 0.71 & 0.92 & 0.80 & 13 \\
\midrule
\multirow{4}{*}{BERT}
 & \multirow{2}{*}{Collaboration} & Positive & 0.94 & 0.98 & 0.96 & 45 \\
 &                                & Negative & 0.50 & 0.25 & 0.33 & 4  \\
 & \multirow{2}{*}{Communication} & Positive & 0.85 & 0.94 & 0.89 & 36 \\
 &                                & Negative & 0.78 & 0.54 & 0.64 & 13 \\
\bottomrule
\end{tabular}
\end{table}

\subsection{Per-Class Results for the Fine-tuned BERT Classifier}
\label{app:bert-per-class}

\begin{figure}[t]
\centering
\includegraphics[width=0.5\textwidth]{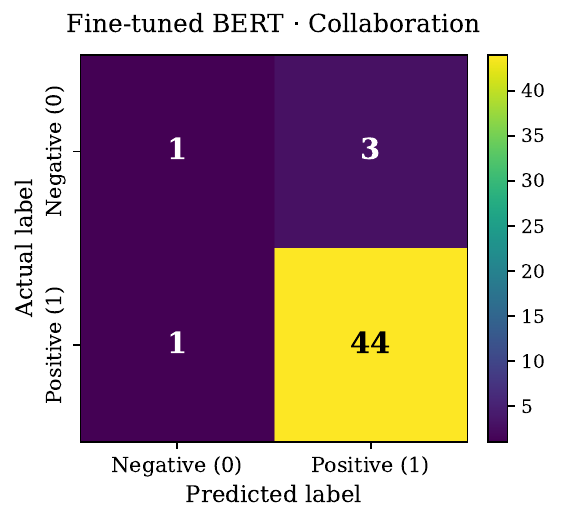}
\caption{Confusion matrices for the fine-tuned BERT classifier on the human-agreed test set -- collaboration.}
\label{confusion-matrices-bert-collaboration}
\end{figure}

For collaboration, the positive class is handled with high fidelity (precision 0.94, recall 0.98, F1 0.96). Forty-four of forty-five positive instances are recovered. The negative class, however, is barely identified (precision of 0.50, recall of 0.25, F1 of 0.33). Only one of the four negative postings is correctly classified, the rest being labeled positive (Fig.~\ref{confusion-matrices-bert-collaboration}). The classifier, therefore, exhibits the textbook signature of a model that has shifted its decision boundary toward the majority class under supervision with limited counterexamples.

\begin{figure}[t]
\centering
\includegraphics[width=0.5\textwidth]{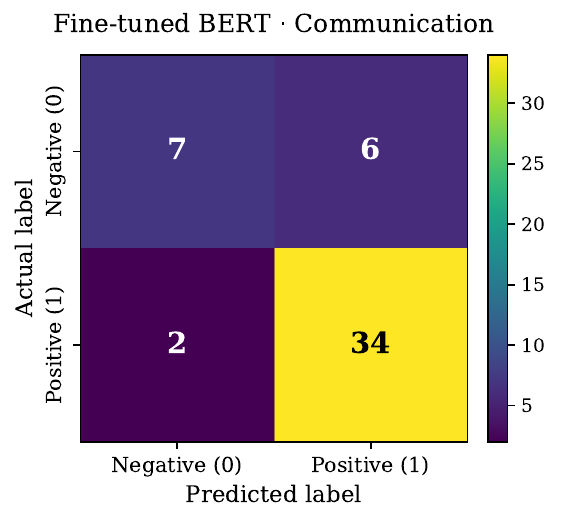}
\caption{Confusion matrices for the fine-tuned BERT classifier on the human-agreed test set -- communication.}
\label{confusion-matrices-bert-communication}
\end{figure}

For communication, the positive class is handled well (precision of 0.85, recall of 0.94, F1 of 0.89; only two positive instances are missed), but performance on the negative class is markedly weaker (precision of 0.78, recall of 0.54, F1 of 0.64). Six of thirteen negative postings are misclassified as positive. The confusion matrix in Fig.~\ref{confusion-matrices-bert-communication} makes this asymmetry visible; the model recognizes explicit cues but struggles to confidently reject postings lacking them.

\begin{acknowledgements}
The project on which this work is based was funded by the Federal Ministry of Education and Research (BMBF) under the funding code 01IS23069.
\end{acknowledgements}

\section*{Conflict of Interest}
The authors declared that they have no known competing interests or personal relationships that could have influenced the work reported in this paper.

\section*{Data Availability}

All artifacts containing the full approach pipelines, datasets, codes, and prompts are available at \url{https://doi.org/10.5281/zenodo.24205893}.

\bibliographystyle{spbasic}
\bibliography{references}

\end{document}